\def\year{2018}\relax
\documentclass[letterpaper]{article} 
\usepackage{aaai18}  
\usepackage{times}  
\usepackage{helvet}  
\usepackage{courier}  
\usepackage{url}  
\usepackage{graphicx}  
\usepackage[compact]{titlesec}
\usepackage{url}            
\usepackage{booktabs}       
\usepackage{amsfonts}       
\usepackage{nicefrac}       
\usepackage{microtype}      
\usepackage{sidecap}
\usepackage{subcaption}
\usepackage{paralist}
\title{AJILE Movement Prediction: \\Multimodal Deep Learning for Natural Human Neural Recordings and Video}

\author{Nancy X.R. Wang$^1$ \and Ali Farhadi$^{1,2}$ \and Rajesh P.N. Rao$^{1,4}$ \and Bingni W. Brunton$^{3}$\\
$^1$Paul Allen School of Computer Science-University of Washington, $^2$Allen Institute for AI, \\
$^3$Department of Biology-University of Washington, $^4$Center for Sensorimotor Neural Engineering (CSNE) \\ \{wangnxr, farhadi, rpnr, bbrunton\}@uw.edu }

\begin{document}
\maketitle

\begin{abstract}

Developing useful interfaces between brains and machines is a grand challenge of neuroengineering.
An effective interface has the capacity to not only interpret neural signals, but \emph{predict} the intentions of the human to perform an action in the near future; prediction is made even more challenging outside well-controlled laboratory experiments.
This paper describes our approach to detect and to predict \emph{natural} human arm movements in the future, a key challenge in brain computer interfacing that has never before been attempted.
We introduce the novel Annotated Joints in Long-term ECoG (AJILE) dataset; 
AJILE includes automatically annotated poses of 7 upper body joints for four human subjects over 670 total hours (more than 72 million frames), along with the corresponding simultaneously acquired intracranial neural recordings.
The size and scope of AJILE greatly exceeds all previous datasets with movements and electrocorticography (ECoG), making it possible to take a deep learning approach to movement prediction.
We propose a multimodal model that combines deep convolutional neural networks (CNN) with long short-term memory (LSTM) blocks, leveraging both ECoG and video modalities.
We demonstrate that our models are able to detect movements and predict future movements up to 800 msec before movement initiation. 
Further, our multimodal movement prediction models exhibit resilience to simulated ablation of input neural signals. 
We believe a multimodal approach to natural neural decoding that takes context into account is critical in advancing bioelectronic technologies and human neuroscience. 
\end{abstract}

\section{Introduction}


Scientists, engineers, and speculative fiction authors have long imagined possible futures when people interact meaningfully with machines directly using thought.
Technologies that interpret brain signals to control robotic and virtual devices have tremendous potential to assist individuals with physical and neurological disabilities, to augment engineered systems integrating humans in the loop, and to enhance one's daily life in an increasingly information-rich world.

In recent years, research in brain-computer interfacing (BCI)~\cite{Wolpaw2012,rao2013} has been very successful in using decoded neural signals to control robotic prostheses and computer software (for instance, \cite{hochberg2012,McMullen2014,Yanagisawa2011}).
Even so, these impressive demonstrations have relied on finely tuned models trained on experimentally derived labeled data acquired in well-controlled laboratory conditions.
Thus, the remarkable feats of neural decoding to mobilize patients who have lost use of their limbs remain untested outside the laboratory. 

One key challenge is how neural decoding may be approached ``in the wild,'' where sources of behavioral and recording variability are significantly larger than what is found in the lab.
Further, neural responses are known to differ between experimental and freely behaving conditions~\cite{Jackson2007}.
A flexible, scalable approach to detect movement and to predict initiation of natural movement would critically enable technologies to foster seamless collaborations between humans and machines.

\begin{figure*}[t]
\includegraphics[width=\textwidth]{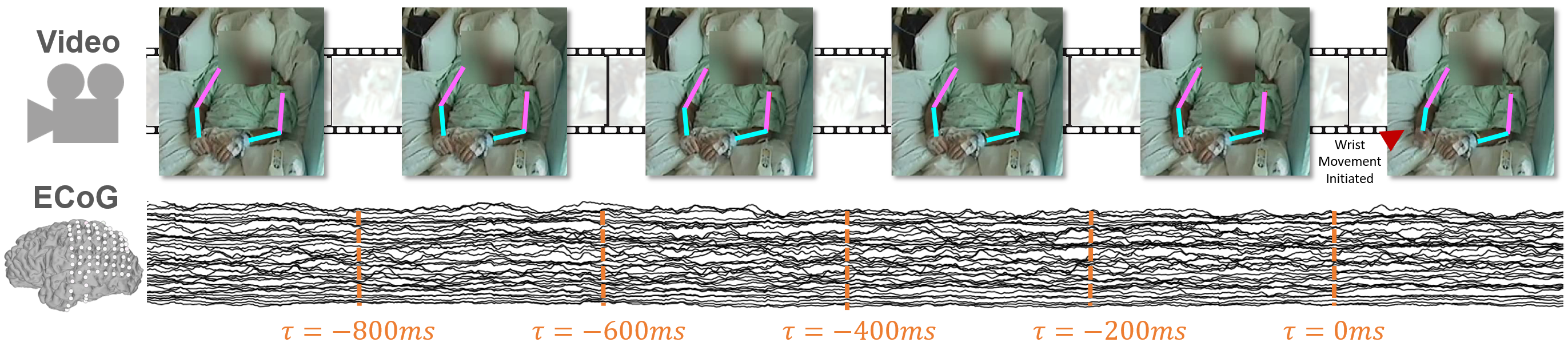} %
\caption{An example of a 1-sec window of multimodal data. 
In each video frame, we show the tracked positions of the upper arm (pink) and forearm (turquoise) for the subject's two arms by the pose recognition algorithm. A right wrist movement was detected in the last frame highlighted. Traces of neural activity acquired by ECoG are shown in black, where deflections represent voltage and a subset of the electrodes are shown stacked vertically.}
\label{fig:samples}
\end{figure*}


In this paper, we present a multimodal deep learning approach that is able to detect whether a subject is initiating a movement and to predict initiation of natural movements in the future.
This paper is the first to develop a deep neural network that models naturalistic ECoG signals. 
The main contributions of the paper are:

\begin{compactitem}
\item We introduce the AJILE dataset of long-term natural neural recordings with corresponding labeled arm poses.

\item Using the AJILE dataset, we show that our proposed multimodal deep neural network can \emph{predict} the intention to move a hand up to 800 msec before the initiation of the movement in naturalistic data, a task never attempted before.


\item Our experimental evaluations show the significance of incorporating context provided by the video, using a deep multimodal model to decode and predict movement intention.

\end{compactitem}

\section{Related work}

\textbf{Human intracranial cortical recordings.} 
Neural signals used to monitor brain activity are acquired by a variety of recording technologies that differ by invasiveness and portability.
One such technology known as electrocorticography (ECoG) for intracranial recordings is particularly attractive for studying naturalistic behavior because of its spatial specificity, temporal resolution, and potential longevity of implantation.
ECoG offers measurements of temporal dynamics inaccessible by functional magnetic resonance imaging (fMRI) and spatial resolution unavailable to extracranial electroencephalography (EEG).
In addition, it is not feasible to use fMRI when one is interested in behaviors over hours and days, especially behaviors that require gross motor movement and meaningful interaction with the surrounding environment.
Cortical surface ECoG is accomplished less invasively than with penetrating electrodes~\cite{Moran2010,Williams2007} and has much greater signal-to-noise ratio than entirely non-invasive techniques such as EEG~\cite{Lal,Ball2009}.

\textbf{Decoding movement.}
Efforts to decode neural activity have been accomplished by training algorithms on tightly controlled experimental data with repeated trials.
Recent examples include decoding arm trajectories~\cite{Nakanishi2013,Wang2013} and finger movements~\cite{Miller2009,Wang2010}.
Decoded neural signals have been used to control robotic arms~\cite{McMullen2014,Yanagisawa2011,Fifer2014} and to construct BCIs~\cite{Wang2013a,Leuthardt2011,Miller2010,Schalk2008}.

\textbf{Analyzing naturalistic brain recordings.}
The lack of ground-truth labels makes analyzing and decoding naturalistic neural recordings especially challenging.
Labels must be obtained by a separate measurement acquired simultaneously with the neural recordings.
Previous studies exploring this idea have decoded natural speech~\cite{Derix2014,Dastjerdi2013} and natural motions of grasping~\cite{Pistohl2012,Ruescher2013}; however, these studies had relied on laborious manual annotations and do not reach the size and comprehensiveness of our AJILE dataset. 
A few studies have made use of some automation to analyze natural data~\cite{wang2016unsupervised,gabriel2016neural}, but none had focused on prediction of future events. 
Our work takes advantage of recent advances in computer vision to annotate a variety of natural data including automated movement estimation~\cite{Wang2015,Poppe2007} and pose recognition~\cite{Toshev2014,Pfister15a}. 

\textbf{Deep neural networks for ECoG/EEG data.}
Deep learning has not been widely applied to ECoG and EEG data, with notable exception in the cases of~\cite{krug2008cnn,wang2013deep,nurse2016decoding,schirrmeister2017deep}.
We are inspired by work in related fields that have made use of multimodal data streams. 
For instance, multimodal networks have been most widely used in the tasks of speaker identification~\cite{ren2016look} and speech recognition~\cite{ngiam2011multimodal}.
Our approach also has similarities to~\cite{Aytar2016}, where visual data was leveraged to derive sound categories from an otherwise unlabeled large dataset.
To our knowledge, there is no previous work combining ECoG with another modality of measurement for developing deep learning models.

\section{Dataset}

\subsection{Long-term, naturalistic neural activity and video}

Our long-term, naturalistic human movement dataset includes week-long continuous multimodal recordings with invasive (ECoG) electrodes, video, and audio.
This opportunistic dataset greatly surpasses all previously analyzed comparable datasets in duration and size.
In total, we have approximately 670 hours of recordings for 4 subjects, which amounts to more than 72 million frames of video and more than 2 billion samples of ECoG at a sampling rate of 1000Hz. 
Importantly, these subjects performed no instructed tasks; instead, they simply did as they wished for the duration of their monitoring in the hospital room.
The variety of natural behaviors observed was rich and complex, including conversations with friends and family, eating, interacting with electronic devices, and sleeping.
Understanding the connection between neural activity and naturalistic behavior presents a great data analytic challenge, in part because of the immense task of annotating such long-term recordings.
At the same time, making sense of this data presents an unique opportunity to shed light on neural function outside the laboratory.
%

\subsection{The AJILE dataset: Annotating joint locations}
In this study, we leveraged the latest innovations in computer vision to train a deep neural network to automatically retrieve the patient pose from each frame.
We used the YOLO~\cite{redmon2016you} framework for subject detection and caffe-heatmap~\cite{Pfister15a} for pose estimation. 
Fig.~\ref{fig:samples} shows a schematic of pixel locations of the head, shoulders, elbows, and wrists as they were extracted from the video frames.
To improve the performance of the standard trained models on our dataset, we acquired custom manual annotations on a small fraction of the video to retrain YOLO and caffe-heatmap. 
A small portion of videos taken from 18 subjects was annotated, with one frame manually labeled approximated once every 2 minutes.
All four subjects in this study were part of the pose estimation training set. 
After over 3000 GPU hours of processing, we extracted locations of 7 upper body joints for over 72 million frames. 
Fig.~S1 shows a validation for the accuracy of these joint locations; our pose estimation is extremely accurate for confidence scores above 0.25, so this threshold was chosen as the cutoff for extraction of natural movement annotations. For a typical patient, approximately half of the frames have a confidence score above 0.25.

We plan to make publicly available this Annotated Joints in Long-term ECoG (AJILE) dataset, at the time of publication of this manuscript.
AJILE includes raw ECoG voltage recordings, electrode locations, and estimated pose in each video frame. 
The pose comprises pixel locations of 7 upper body joints, along with an estimated confidence value for each joint.
AJILE does not include raw video recordings, a restriction due to patient privacy.

\subsection{Extracting initiation of natural movements from AJILE}
To define movement from estimated poses, we focused on wrist movements of the arm contralateral to the electrode array implant. 
We first smoothed the joint location results from AJILE using a Savitzky-Golay filter with a 21-frame window. 
A movement initiation was defined as when movement of the wrist joint averaged over 5 consecutive video frames exceeded an average of 1 pixel per frame, and less than 0.5 pixels of movement was detected when averaged over the previous 10 frames.
Times of no movement were selected when there was less than 0.5 pixels of movement in all joints averaged over 30 frames before and after the time point in question. 

A small portion of automated movement detection was validated with manual annotations. 
We found the inter-rater reliability to be 95.3\% and the overall accuracy against the raters are 86.9\% and 84.2\%. 
After the automated process, the data was curated to discard obviously inaccurate labels.
For example, we removed samples where there was another person moving in front of or obstructing the arm, as well as samples during sleep, since neural patterns are known to be drastically different between sleep and wake.
Table~\ref{tab:trainingset} summarizes the number of instances of movement initiation in the dataset for each of the four subjects.
The training data includes movements from days 2 to 5 of the clinical monitoring, and day 6 or 7 was used for testing.
Each train and test dataset was balanced so that they contained roughly equal numbers of movement and no movement samples. 

\subsection{Data preprocessing}

All ECoG recordings were bandpass filtered between 10 and 200Hz. 
For the neural network models, a 1-second window of high-dimensional time-series data was used as the input (shown schematically in Fig. 1). Each 1-second window of recording for each electrode was normalized to the mean and standard deviation of its 3-second neighborhood that does not contain any times of movement, then broken into a sequence of five 200 msec chunks. These chunks were used as inputs to the neural network. 

For our multimodal model, each 200 msec chunk of data was associated with one video frame, which was extracted from the middle of the 200 msec time window. Video frames were resized from $640\times480$ down to $341\times256$. During training, the ECoG data was augmented with noisy perturbations 25\% of the time using gaussian random noise of standard deviation 0.001 and temporal shifting of up to 100 msec in either direction. All video frames during training were randomly cropped into $224\times224$ images for input into the networks. During testing, the images were always cropped at the center of the frame. 

In all training schemes, each subject's data is trained and tested separately and independently. Combining data across subjects was not possibly because of large differences in electrode coverage. 

\begin{table}[h]
\centering
\begin{tabular}{lrrrr}
\toprule
& S1   & S2   & S3   & S4   \\ 
\midrule
Train & 1560 & 2002 & 4587 & 3490 \\ 
Test  & 313  & 575  & 1952 & 193  \\ 
\bottomrule
\end{tabular}
\caption{Number of samples in the dataset for each subject. The test set was chosen to be on a day of recording different from the training set; variations are due to the activeness of each subject.}
\label{tab:trainingset}
\end{table}

\subsection{Clinical data collection details}

The subjects in our dataset were patients undergoing pre-surgical clinical epilepsy monitoring.
The study was approved by our institute's human subject division; all four (4) subjects gave their informed consent and all methods were carried out in accordance with the approved guidelines. 
Electrode placement and duration of each subject's recording were determined solely based on clinical needs. 
Each subject had 80--94 ECoG electrodes implanted subdurally, which is to say, directly on the brain under the skull and dura, a tough membrane surrounding the brain. 
S1, S2, and S3 had electrode implants in the left brain hemisphere; S4 was implanted in the right hemisphere. 
The electrodes are arranged as grids of $8\times8$, $8\times4$, $8\times2$ or strips of $1\times4$, $1\times6$, $1\times8$. 
Electrode grids were constructed of 3-mm-diameter platinum pads spaced at 1 cm center-to-center and embedded in silastic (AdTech). 
Electrodes that experienced failure during the subject's recording were rejected from the dataset.
Fig. S2-S5 show the electrode placements of each subject. 
All subjects had between 6 and 7 days of continuous monitoring with video, audio and ECoG recordings. 
Video and audio were recorded simultaneously with the ECoG signals and continuously throughout the subjects' clinical monitoring. 
Generally, video was centered on the subject with family members or staff occasionally entering the scene. 
The video was recorded at 30 frames per second at a resolution of $640\times480$ pixels. The camera was sometimes adjusted throughout the day by hospital staff. 
For example, the camera may be moved during bed pan changes and returned to the subject afterwards, but not always to exactly the same position. 
Fig.~\ref{fig:samples} shows example video frames from one subject; the face was blurred to protect their privacy. 

\section{Prediction of movement in the future}
\subsection{Defining the problem}

The task we address in this paper is the prediction of spontaneously generated arm movements in the future.
A viable solution to this problem is critical for the application of brain-computer interfacing ``in the wild,'' where a model must be able to tolerate significant noise and variability in natural, uncontrolled conditions. 
To our knowledge, this task has never before been attempted.


\subsection{Our approach}

\begin{figure}[h!]
\begin{subfigure}{0.47\textwidth}
\caption{Multimodal neural network schematic}
\resizebox{\linewidth}{!}{%
\includegraphics{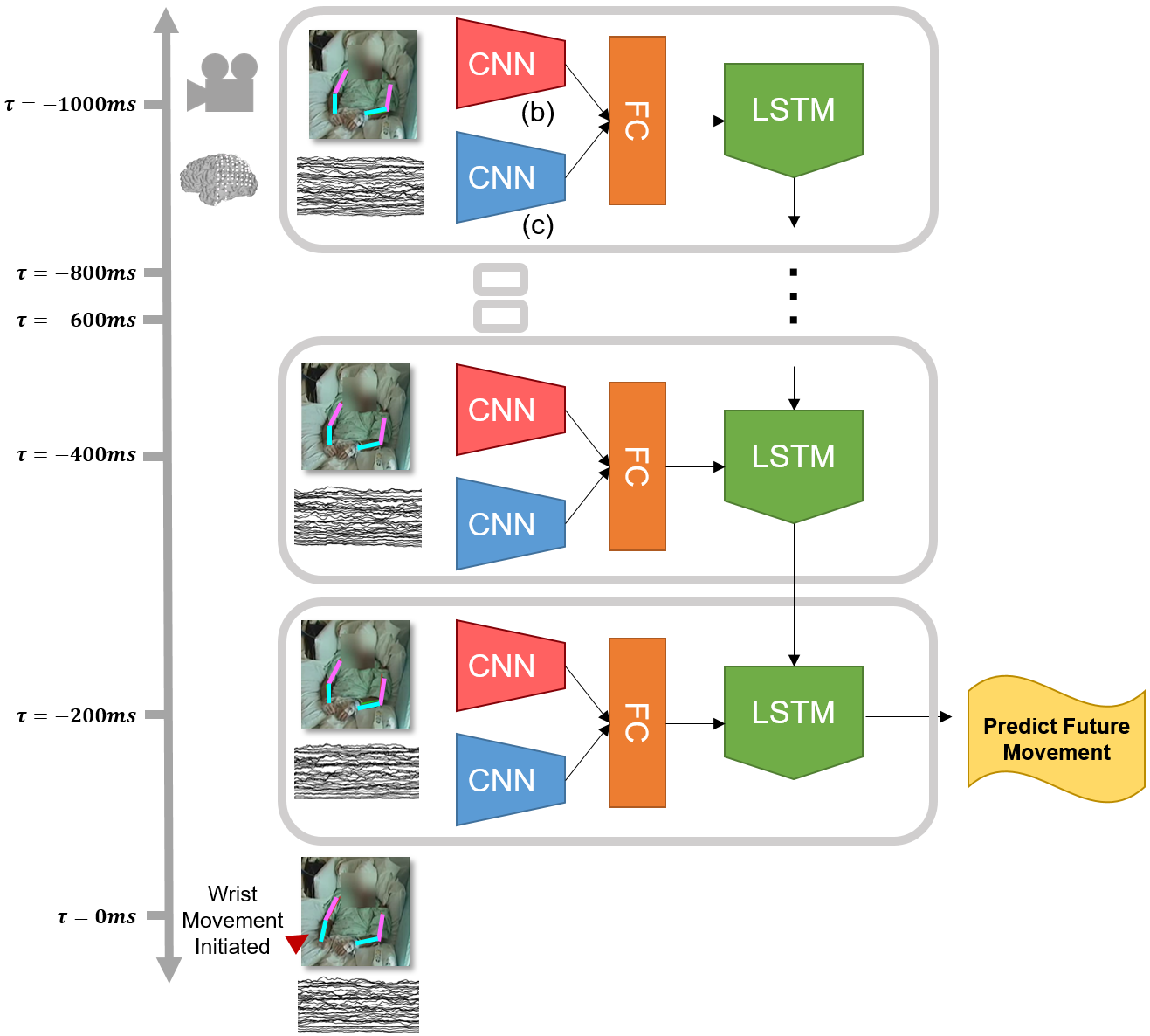}
}
\end{subfigure}

\begin{subfigure}{0.47\textwidth}
\caption{Video CNN architecture}
\resizebox{\linewidth}{!}{%
\includegraphics{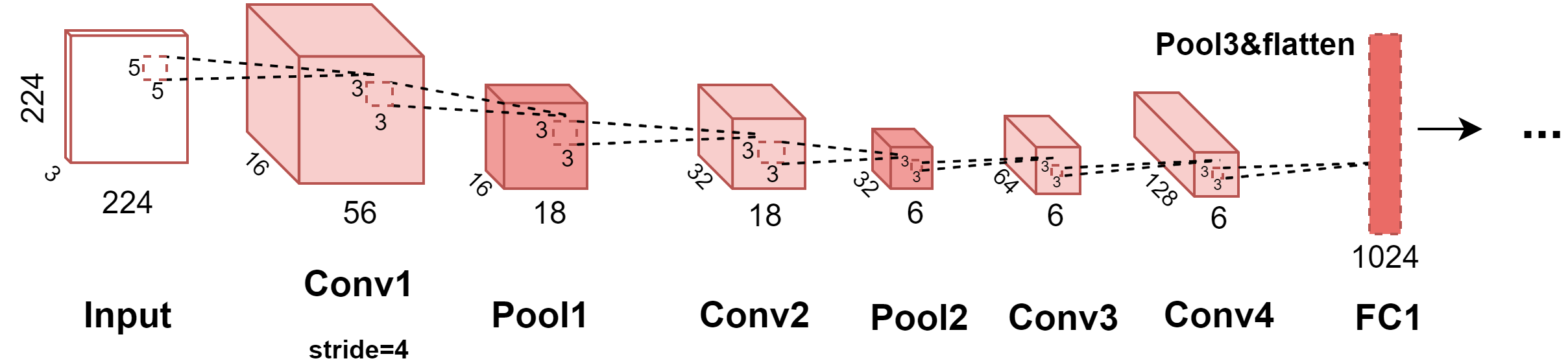} %
}
\end{subfigure}

\begin{subfigure}{0.47\textwidth}
\caption{ECoG CNN architecture}
\resizebox{\linewidth}{!}{%
\includegraphics{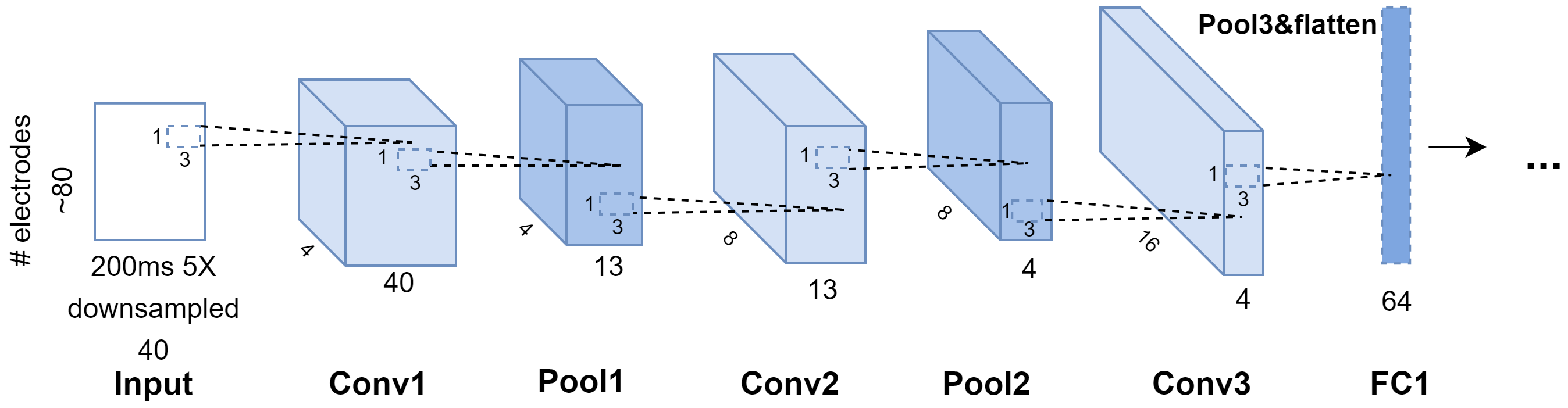}
}
\end{subfigure}

\caption{A schematic of the multimodal neural network architecture for the prediction of future hand movements using ECoG and video frame data. The ECoG data was separated into five 200-msec chunks. Each ECoG chunk and one frame at the center of the 200-msec time period was extracted from the video to use in the input sequence for the CNN/LSTM neural network.
\label{fig:architecture}}
\end{figure}

Our approach to this prediction problem is inspired by recent advances in multimodal deep neural networks.
We formulated a movement initiation classification problem using training and test data extracted from the AJILE dataset (Table~\ref{tab:trainingset}).
We reasoned that such a flexible framework would adapt to the dynamic environment in the data, synthesizing cues from direct recording of brain activity and contextual information provided by the video.
Our multimodal model (Fig.~\ref{fig:architecture}) comprises two parallel towers, one 3-layer 1D CNN for ECoG and one 4-layer 2D CNN for video inputs, which are then merged with a fully connected layer followed by a LSTM layer of 20 units.
Input data are fed into the CNN in 5 sequential chunks that includes one second of recordings in total. 

Although neural activity is the ultimate director of one's future actions, ECoG is a very incomplete sampling of the brain, so we believe information from the video adds context that may improve accuracy in the prediction problem.
In addition, multimodal information should make the model more robust to noise and variability than with a single modality alone. 
ECoG and video data are both sequential by nature, so we developed a sequential model to match. 
However, we know from extensive literature analyzing ECoG signals that power-frequency features are usually more informative than raw voltage, so the convolutional layers act as feature extraction before the sequential layer.
Finally, we decided to fuse ECoG and video towers in a late fusion model, hypothesizing that features that we can extract using the CNN would be different for each modality and should have different filtering sizes and layer structure.

\begin{table*}[h]
\centering
\caption{Multimodal vs. ECoG only vs. Video only
\label{tab:modalities}
}
\begin{tabular}{@{}rrrrrrrrrrrr@{}}
\toprule
& \multicolumn{3}{c}{Multimodal} &~& \multicolumn{3}{c}{ECoG only} &~& \multicolumn{3}{c}{Video only} \\
& Pred-b & Pred & Det && Pred-b & Pred & Det && Pred-b & Pred & Det \\
\cmidrule{2-4} \cmidrule{6-8} \cmidrule{10-12} 
\textbf{S1}        & 76.8  & 81.9 & 87.0 && 66.8 & 58.6 & 66.4 && 86.7 & 75.9 & 81.3\\ 
\textbf{S2}        & 59.4  & 61.7 & 61.5 && 66.7 & 65.7 & 64.6 && 58.0 & 49.8 & 52.1\\ 
\textbf{S3}        & 67.9  & 71.2 & 79.8 && 66.1 & 68.3 & 83.2 && 65.7 & 64.0 & 65.4\\  
\textbf{S4}        & 66.1  & 62.0 & 62.5 && 56.8 & 54.7 & 57.3 && 49.0 & 54.7 & 57.8\\ 
\cmidrule{2-4} \cmidrule{6-8} \cmidrule{10-12} 

\textbf{Average}   & \textbf{67.6}  & \textbf{69.2} & \textbf{72.7} && 64.1 & 61.8 & 67.9 && 64.9 & 61.1 & 64.2 \\
\bottomrule
\end{tabular}
\end{table*}

\begin{table*}[h]
\centering
\caption{Late Fusion vs. Early Fusion vs. Naive Averaging
\label{tab:earlyvlate}
}
\begin{tabular}{@{}rrrrrrrrrrrr@{}}
\toprule
& \multicolumn{3}{c}{Late Fusion} &~& \multicolumn{3}{c}{Early Fusion} &~& \multicolumn{3}{c}{Naive Averaging} \\
& Pred-b & Pred & Det && Pred-b & Pred & Det && Pred-b & Pred & Det \\
\cmidrule{2-4} \cmidrule{6-8} \cmidrule{10-12} 
\textbf{S1}        & 76.8  & 81.9 & 87.0 && 82.0 & 67.7 & 78.1 && 85.9 & 80.4 & 85.4\\ 
\textbf{S2}        & 59.4  & 61.7 & 61.5 && 56.1 & 51.2 & 60.6 && 62.5 & 57.3 & 58.0  \\ 
\textbf{S3}        & 67.9  & 71.2 & 79.8 && 52.2 & 55.7 & 79.4 && 69.1 & 68.4 & 72.4\\  
\textbf{S4}        & 66.1  & 62.0 & 62.5 && 53.1 & 69.3 & 55.2 && 50.0 & 60.9 & 55.7\\ 
\cmidrule{2-4} \cmidrule{6-8} \cmidrule{10-12} 

\textbf{Average}   & \textbf{67.6}  & \textbf{69.2} & \textbf{72.7} && 60.9 & 61.0 & 68.3 && 66.9 & 66.8 & 67.9 \\
\bottomrule
\end{tabular}
\end{table*}

\section{Experiments}

Our proposed multimodal neural network is able to detect and predict future movements.
We present results for models trained on three different timing conditions: detection (Det), prediction (Pred), and further back prediction (Pred-b).
For detection, the 1-sec window of data was centered at the time of initiation of movement (i.e., 500 msec before and 500 msec after).
For prediction, data was taken 1300 msec to 300 msec before the movement, so that the initiation itself was not included in the data.
For further back prediction, data was taken 1800 msec to 800 msec before the movement.
Table~\ref{tab:modalities} shows that, on average across four subjects, the multimodal model performs well above chance (50\%) and out-performs similar models that use a single modality of data in every timing condition.

In the remainder of this section, we describe our analyses of the multimodal model and demonstrate that our proposed approach has advantages over similar, related models.
These analyses highlight the significance of combining context from video with direct neural recordings to enhance movement prediction.
Further, we report results from synthetic electrode ablation experiments to evaluate the resilience of the multimodal neural network model.

\subsection{Analyses of the multimodal model}

\textbf{Optimal Fusion Point.} Our multimodal model uses video and ECoG streams of data after each stream has been separately fed through their respective CNN's (Fig.~\ref{fig:architecture}).
We refer to this fusion scheme as ``late fusion.''
To compare this strategy with other potential points of fusion for the video and ECoG data, we trained an alternative ``early fusion'' model that stacks ECoG with video directly in the input, resizing the image as needed.
In addition, we compare with a ``naive average'' model that averages the final sigmoid outputs of the ECoG only and video only models.
Table~\ref{tab:earlyvlate} shows that ``late fusion'' outperforms both of the other schemes.
Early fusion likely suffers from forcing the two modalities to have the same CNN architecture when many hyperparameters (such as filter size) should be quite different for the very different types of data. 
The improvement in accuracy as compared to naive averaging suggests that the fully connected layers and LSTM after the merge layer are learning aspects that are multimodal in nature, beyond a simple combination of probabilities from each modality.

\textbf{Importance of CNN.} Since ECoG is a sequential data type, we investigated the potential of a purely sequential model for the classification task. 
As shown in Table~\ref{tab:lstm}, the LSTM-only model performs at around chance. 
This observation shows the importance of feature extraction from the CNN layers before classification.

\textbf{ECoG filter dimensions.} Each subject in the dataset has at least one large grid of electrodes that is $8\times8$ in shape. 
Since this electrode geometry is known, we investigated to the potential of using a 3-dimensional convolutional filter on the ECoG data to take advantage of neighboring electrode positions.
In direct comparison with the 1D filters used in our proposed model, which filters the ECoG data only in time, the 3D convolutional filters do not perform as well(Table~\ref{tab:3dconv}).
In this comparison, we removed the LSTM portion of the models to more directly compare the CNN filter schemes. 
We speculate that the time domain is more informative for predicting hand movement than the spatial domain. 
When our convolutional filters and pooling involve both space and time, this may be reducing the amount of information that can be obtained from the time domain. 

\textbf{Comparison to traditional baseline.} To compare our deep learning approach to models more conventionally used to analyze ECoG data, we developed a baseline model using a linear SVM classifier based on power spectral features (see for instance~\cite{shenoy2008,Yanagisawa2011}).
The power spectral features in two frequency bands (10--30 Hz and 70--100 Hz) were extracted using short-time Fourier transform using non-overlapping 1-sec windows. 
Model selection of the baseline model was performed using a validation set drawn from the training days. 
Table~\ref{tab:svm} shows our deep learning ECoG only model outperforms the traditional spectral feature based SVM model.

\textbf{Resilience of models after virtual electrode ablation.} An important feature of movement detection and prediction models is robustness to disturbances. Here we investigated one type of robustness, namely the resilience of each model to a catastrophic electrode failure. This type of failure, when an electrode becomes entirely not functional, is not uncommon in real-life; the point of failure may be due to the electrode/amplifier interface, the wire connection, or movement/scarring of the brain tissue.

To simulate electrode failure, we systematically ablated each individual electrode in turn, substituting its true signal with a constant set to its mean value over time. Thus, we map each electrode's importance in making a detection or prediction by the impact of its ablation on the overall model accuracy. These maps also allow us to directly compare key electrode locations with known cortical maps from the human neuroscience literature.

\begin{table}[h]
\centering
\caption{Conv + LSTM vs. LSTM only
\label{tab:lstm}
}
\resizebox{\linewidth}{!}{%
\begin{tabular}{@{}rrrrrrrr@{}}
\toprule
& \multicolumn{3}{c}{Conv + LSTM} &~& \multicolumn{3}{c}{LSTM only} \\
& Pred-b & Pred & Det && Pred-b & Pred & Det \\
\cmidrule{2-4} \cmidrule{6-8} 
\textbf{S1}        & 66.8  & 58.6 & 66.4 && 48.2 & 49.3 & 51.2\\ 
\textbf{S2}        & 66.7  & 65.7 & 64.6 && 48.8 & 53.5 & 50.0\\ 
\textbf{S3}        & 66.1  & 68.3 & 83.2 && 51.6 & 50.7 & 56.8\\  
\textbf{S4}        & 56.8  & 54.7 & 57.3 && 47.9 & 54.2 & 46.9\\ 
\cmidrule{2-4} \cmidrule{6-8} 

\textbf{Average}   & \textbf{64.1}  & \textbf{61.8} & \textbf{67.9} && 49.1 & 51.9 & 51.2\\
\bottomrule
\end{tabular}
}
\end{table}

\begin{table}[h]
\centering
\caption{1D convolutional filters vs. 3D convolutional filters
\label{tab:3dconv}
}
\resizebox{\linewidth}{!}{%
\begin{tabular}{@{}rrrrrrrr@{}}
\toprule
& \multicolumn{3}{c}{1D Conv} &~& \multicolumn{3}{c}{3D Conv} \\
& Pred-b & Pred & Det && Pred-b & Pred & Det \\
\cmidrule{2-4} \cmidrule{6-8} 
\textbf{S1}        & 58.4  & 60.6 & 66.2 && 51.2 & 45.4 & 62.6\\ 
\textbf{S2}        & 63.4  & 65.9 & 69.9 && 57.5 & 64.5 & 63.8\\ 
\textbf{S3}        & 64.8  & 66.3 & 79.8 && 66.6 & 70.7 & 77.9\\  
\textbf{S4}        & 52.6  & 55.7 & 65.1 && 47.4 & 48.4 & 55.7\\ 
\cmidrule{2-4} \cmidrule{6-8} 

\textbf{Average}   & \textbf{59.8}  & \textbf{62.1} & \textbf{70.3} && 55.7 & 57.3 & 65.0\\
\bottomrule
\end{tabular}
}
\end{table}

\begin{table}[h]
\centering
\caption{ECoG deep model vs. SVM with spectral features
\label{tab:svm}
}
\resizebox{\linewidth}{!}{%
\begin{tabular}{@{}rrrrrrrr@{}}
\toprule
& \multicolumn{3}{c}{Deep} &~& \multicolumn{3}{c}{Traditional} \\
& Pred-b & Pred & Det && Pred-b & Pred & Det \\
\cmidrule{2-4} \cmidrule{6-8} 
\textbf{S1}        & 58.4  & 60.6 & 66.2 && 49.5 & 52.6 & 63.2\\ 
\textbf{S2}        & 63.4  & 65.9 & 69.9 && 62.7 & 68.8 & 67.4\\ 
\textbf{S3}        & 64.8  & 66.3 & 79.8 && 50.2 & 50.2 & 62.4\\  
\textbf{S4}        & 52.6  & 55.7 & 65.1 && 53.1 & 51.0 & 50.0\\ 
\cmidrule{2-4} \cmidrule{6-8} 

\textbf{Average}   & \textbf{59.8}  & \textbf{62.1} & \textbf{70.3} && 53.9 & 55.7 & 60.8\\
\bottomrule
\end{tabular}
}
\end{table}

\begin{figure}

\resizebox{\linewidth}{!}{%
\includegraphics{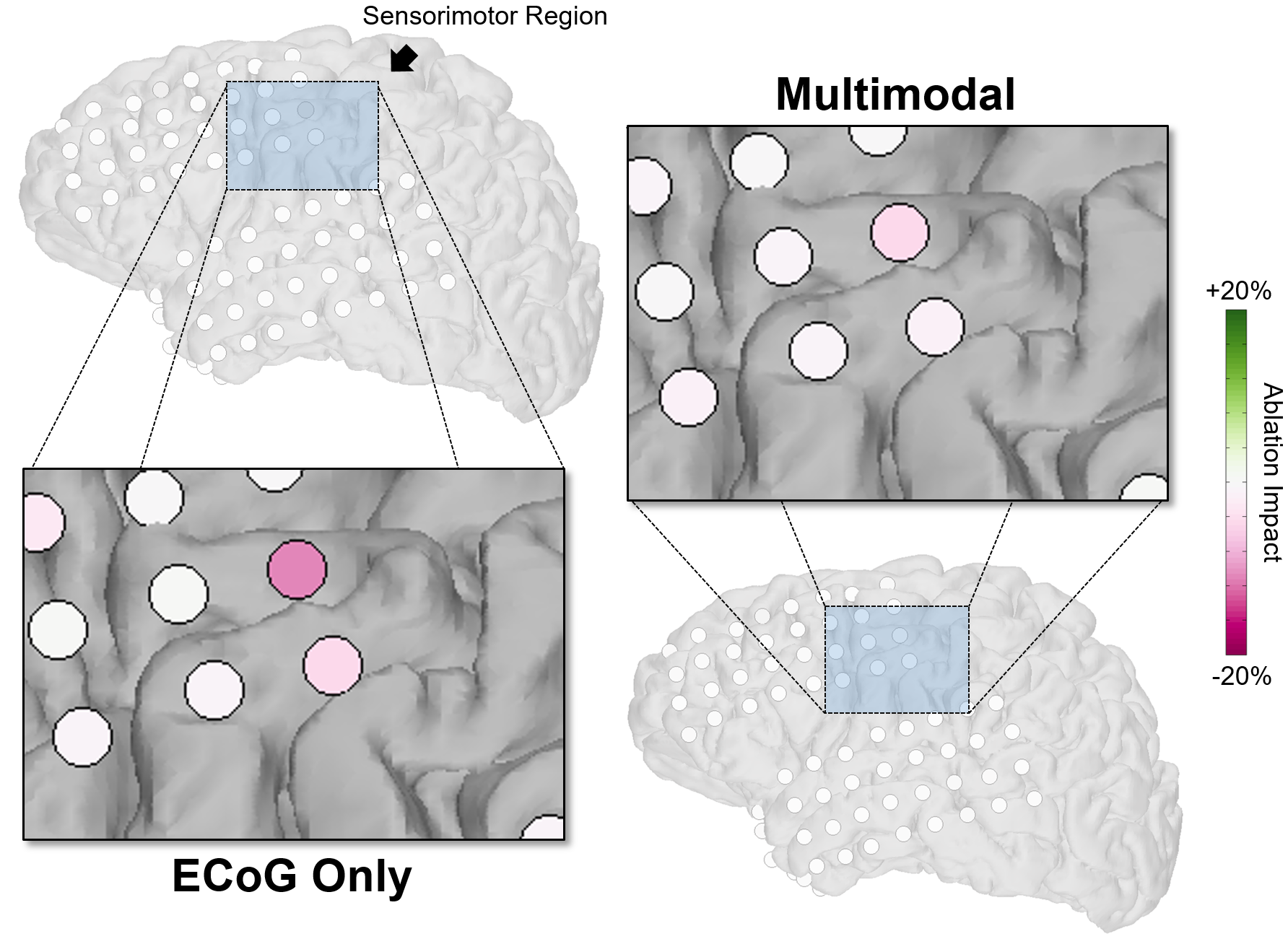} %
}
\caption{Effects of simulated single electrode ablation in the movement detection task for S2 show that the most important electrodes are found in sensorimotor regions. The colormap represents the change in accuracy of the model due to ablation, compared to the intact, original model.
The multimodal model is generally more resilient to single electrode ablations.}
\label{fig:ablate}
\end{figure}

Fig.~\ref{fig:ablate} shows ablation maps for one subject (S2) for the detection task, comparing the resilience of the ECoG only model and the multimodal model. The most important electrode for this subject is in the cortical area corresponding to sensorimotor function. Ablation maps for all subjects and experiments are shown in Fig. S2--S5. The ablation analysis revealed that the most important electrodes for the ECoG only model were those in sensorimotor regions, prefrontal regions (implicated in motor planning), and speech regions (e.g. Broca's and Wernicke's areas), likely explained by co-occurrence of speech and movement in this natural dataset. 

Our multimodal model, on the other hand, was less impacted by ablation of single electrodes when compared to ECoG only (Fig.~\ref{fig:ablate} and Fig. S2--S5). Table~\ref{tab:ablation_score} shows detailed comparisons of intact vs. ablation experiments. In the worst-case single-electrode ablated experiments, the decrease in accuracy of ECoG model after ablation is always larger than the decrease for the multimodal model except in one case. 

In an additional experiment, we ablated all electrodes in the multimodal model, and the accuracy dropped to chance levels. This all-ablation experiment confirms that the multimodal network was not simply ignoring the ECoG input. Instead, the video was able to alleviate dependence on individual electrodes, resulting in a more robust multimodal model.

\begin{table*}[]
\centering
\caption{Ablation resilience of ECoG only vs. multimodal. \textbf{Bold} indicates the model with the higher accuracy post-ablation.}
\resizebox{\linewidth}{!}{%

\begin{tabular}{@{}rrrrrrrrrrrrrrrr@{}}
\toprule
& \multicolumn{3}{c}{S1} &~& \multicolumn{3}{c}{S2} &~& \multicolumn{3}{c}{S3} &~& \multicolumn{3}{c}{S4}\\ 
\cmidrule{2-4} \cmidrule{6-8} \cmidrule{10-12} \cmidrule{14-16} 
& Orig&Ablate&Diff &&Orig&Ablate&Diff &&Orig&Ablate&Diff & &Orig&Ablate&Diff \\ 
\midrule
\textbf{Detect} & & & & & & & & & & & &  \\\
ECoG only & 66.4 & 60.7 & 5.7 && 64.6 & \textbf{62.0} & 2.6 && 83.2 & 56.3 & 26.9 && 57.3 & 53.1 & 4.2\\ 
Multimodal  & 87.0 & \textbf{86.2} & 0.8 && 61.5 & 59.7 & 1.8 && 79.8 & \textbf{78.0} & 1.8 && 62.5 &  \textbf{59.9} &2.6 \\ 
\textbf{Pred} & & & & & & & & & & & & \\ 
ECoG only & 58.6 & 55.8 & 2.8 && 65.7 & 55.6 & 10.1 && 68.3 & 58.7 & 9.6 && 54.7 & 49.0 & 5.7 \\ 
Multimodal & 81.9 & \textbf{78.6} & 3.3 && 61.7 & \textbf{57.5} & 4.2 && 71.2 & \textbf{68.2} & 3.0 && 62.0 & \textbf{57.8} & 4.2 \\ 
\textbf{Pred-b} & & & & & & & & & & & & \\\
ECoG only & 66.8 & 63.0 & 3.8 && 66.7 & \textbf{57.3} & 9.4 && 66.1 &61.2 & 4.9 && 62.0 & 41.7 & 15.1 \\ 
Multimodal & 76.8 & \textbf{74.1} & 2.7 && 59.4 & 56.3 & 3.1 && 67.9 & \textbf{63.9} & 4.0 && 66.1&\textbf{62.5} & 3.6 \\ 
\bottomrule
\end{tabular}%
}
\label{tab:ablation_score}
\end{table*}

\subsection{Implementation details}

We implemented our networks in Tensorflow~\cite{tensorflow2015-whitepaper} with the Keras~\cite{chollet2015keras} module. 
We used a stochastic gradient descent (sgd) optimizer with a learning rate of 0.001, momentum term of 0.9, and decay factor of 0.9 for all experiments. 
The batch size was 24 in order to ensure that the multimodal network would fit in memory. 
The initial weights were generated by the glorot uniform distribution. 
After every convolution, we used rectified linear activation units (ReLU). 
Dropout of 0.5 was applied after each fully connected layer. 
Each network was trained for 200 iterations with an early stopping criteria. 
Each training procedure was run three times, because on some runs, poor initial weights led to very poor final results.
The final model was selected as the model with the best accuracy out of these three runs, as assessed on a validation set randomly sampled from the training days.
Optimization speed varied for different subject sets and experiments but typically took a few hours on a TESLA X (Pascal) GPU.


\section{Discussion}

In this paper, we introduced the AJILE dataset, which contains over 670 hours (over 72 million frames) of total continuous naturalistic human ECoG data with corresponding upper body joint locations. 
AJILE greatly surpasses in scope and size all previous datasets of neural recordings of human movements, allowing deep learning approaches to be applied to neural decoding problems. 
The dataset will be released to coincide with this paper's publication. 
We also presented the first model that successfully predicts future movement from natural human ECoG data. 
The ECoG-video multimodal deep neural network models show improved accuracy and robustness beyond using each modality alone. 


The current work predicts the initiation of movement of the contralateral hand. This approach can be extended to predict and regress the locations of multiple joints for a more detailed reconstruction of future movements. 
Because of significant variation across individual subjects, we believe that deep learning on large quantities of raw data is a more scalable and sustainable approach than models built on hand-crafted features.

\subsection{Visualizing Filters}

\begin{figure}[t]
\centering
\resizebox{0.95\linewidth}{!}{%
\includegraphics{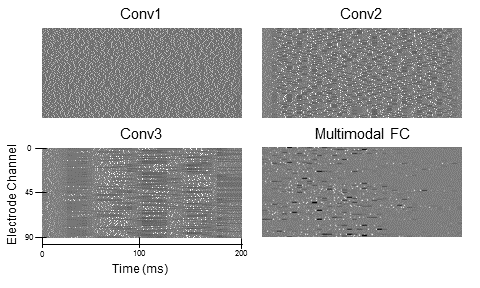} %
}
\caption{Conv1, Conv2 and Conv3 represent input that maximizes activation for sample units across the first three convolutional layers in S1 ECoG only model. Multimodal FC is computed from the S1 multimodal network's first FC. The general pattern is that deeper layers of the network have more unique maximally desired inputs in electrode space and time.
}
\label{fig:weights}
\end{figure}

To investigate features extracted by layers of the neural network models, we used gradient based input optimization, visualizing ECoG inputs that maximally activated filters at various layers. 
Fig.~\ref{fig:weights} shows a few example filters from different parts of the neural network, and we make the general observation that features acquire more distinct structure at deeper layers, which is consistent with what has been described in the image realm. 
In addition, earlier convolutional neural network units tend to show a preference for distinct temporal frequencies in the signal, resembling features represented in a Fourier basis. 
Deeper network layers tend to prefer more complex temporal features, with dynamic frequencies across space and time. 

\subsection{Implications and connections to neuroscience} 

Our approach builds custom models tailored to individuals using only raw data, adapting to variations such as individual electrode placement without expert intervention.
Dissecting these models revealed several observations that have direct connections to human neuroscience.
First, the virtual electrode ablation studies revealed the most important electrode locations lie in an area of cortex known to be sensorimotor cortex.
Second, the convolutional filters learned in the ECoG tower have features similar to Fourier bases, which are by far the most common approach to analyzing ECoG data in neuroscience.
Moreover, our deep neural networks, especially at deeper layers, learn more complex features that are dynamic in space and time, suggesting that the investigation of our models may uncover novel and surprising patterns in neural activity underlying naturalistic movements. 

\subsection{Acknowledgements}

We thank Maya Felten and Ryan Shean for annotation. We also thank Dr. Jeffery Ojemann and Nile Wilson for aiding in data collection as well as research discussions. This research was support by the Washington Research Foundation (WRF), Moore Foundation, Sloan Foundation and the National Science Foundation (NSF) award 1630178 and EEC-1028725.



\small
\bibliographystyle{aaai}
\bibliography{references}

\end{document}


\addtolength{\topmargin}{0.5in}





\begin{figure}[h!]
\centering
\includegraphics[scale=0.6]{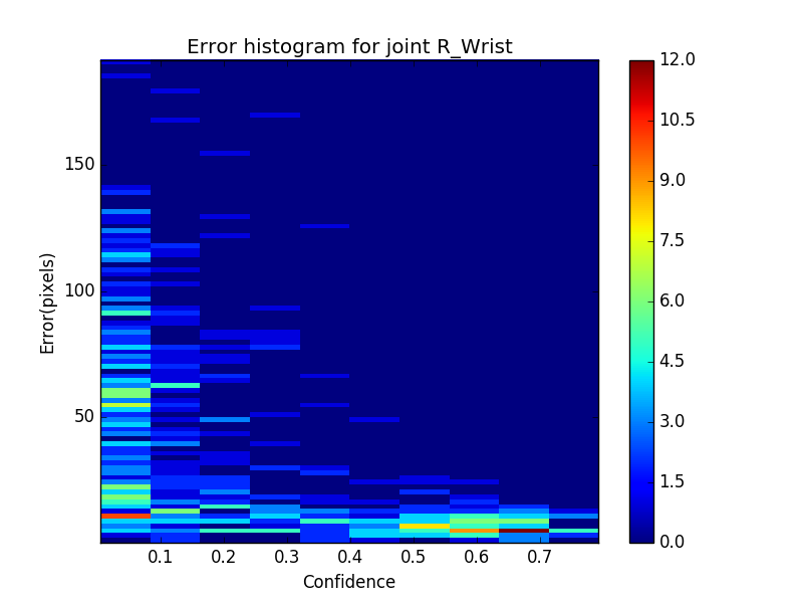}
\includegraphics[scale=0.6]{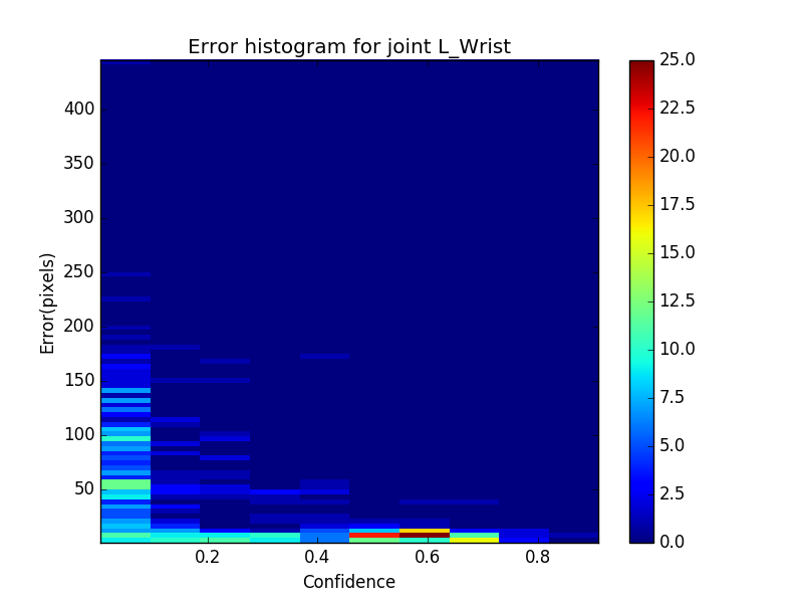}
\caption*{Figure S1: Histograms of accuracy on test data from multiple subjects (not necessarily including subjects in this study) show that when confidence is above 0.25, almost all test instances are quite accurate. For scale, a wrist on the video is approximately 25 pixels wide. Top: Right wrist. Bottom: Left wrist. }
\label{fig:hist}
\end{figure}

\newpage

\centering
\includegraphics[scale = 0.5, angle =90]{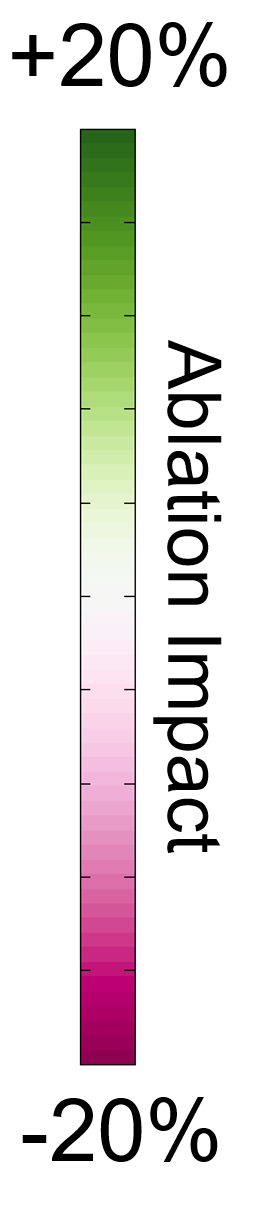}
\begin{table}[h!]
  \centering
  \begin{adjustbox}{max width=\textwidth}
  \begin{tabular}{  c  c  c  }
    
     & ECoG NN & Multimodal NN \\ 
    Pred\_back &
    \begin{minipage}{.55\textwidth}
      \includegraphics[width=\linewidth]{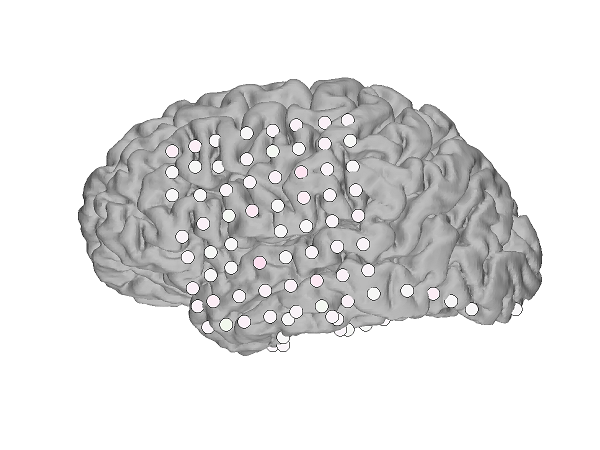}
    \end{minipage}
    & 
    \begin{minipage}{.55\textwidth}
      \includegraphics[width=\linewidth ]{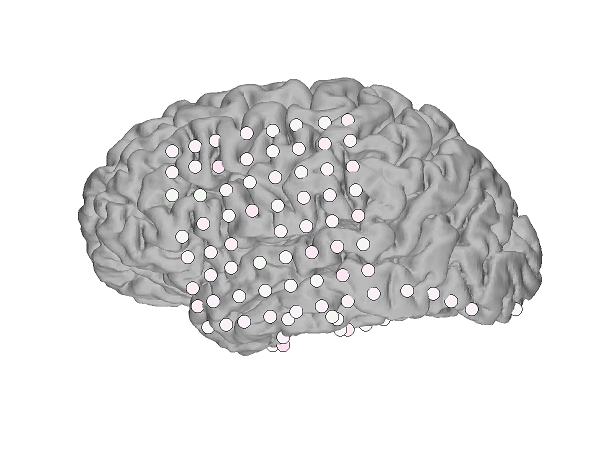}
    \end{minipage} \\ 
    
        Pred & 
    \begin{minipage}{.55\textwidth}
      \includegraphics[width=\linewidth]{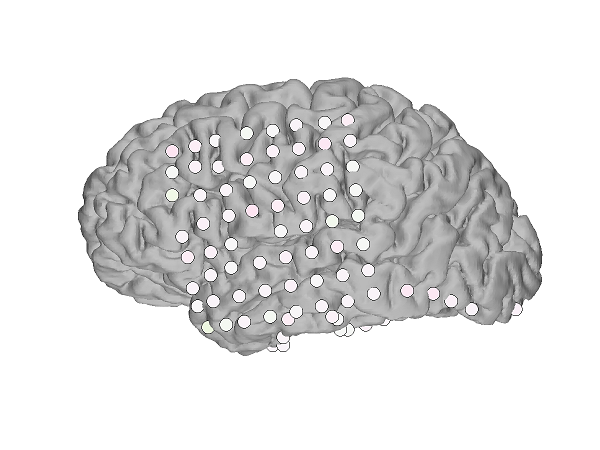}
    \end{minipage}
    &
    \begin{minipage}{.55\textwidth}
      \includegraphics[width=\linewidth]{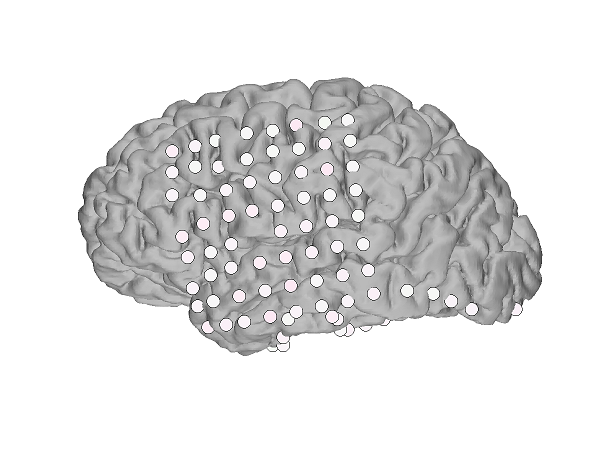}
    \end{minipage} \\ 
    
        Decode &
    \begin{minipage}{.55\textwidth}
      \includegraphics[width=\linewidth]{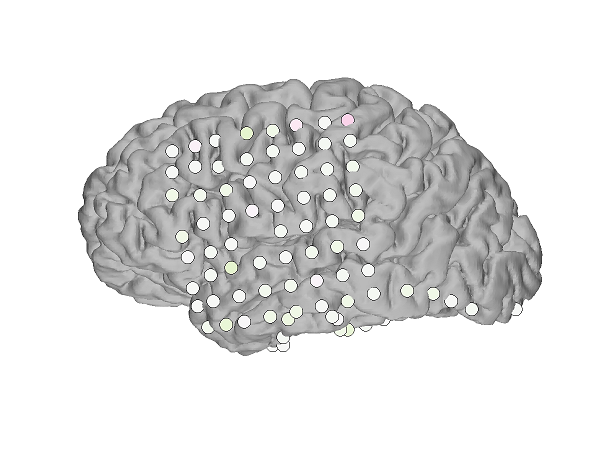}
    \end{minipage}
    &
    \begin{minipage}{.55\textwidth}
      \includegraphics[width=\linewidth]{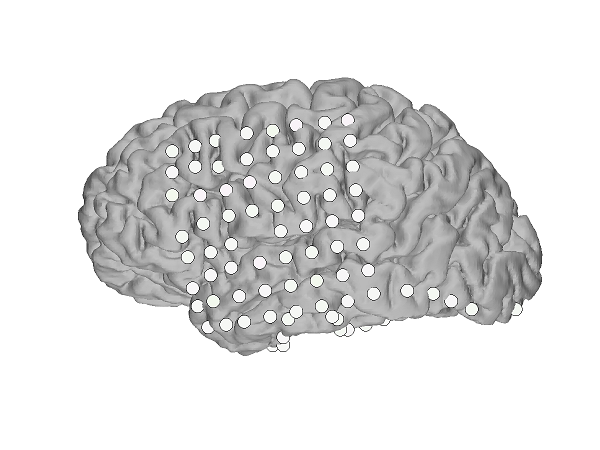}
    \end{minipage} \\ 
  \end{tabular}
  \end{adjustbox}
  \caption*{Figure S2: Subject 1 ablation map. Ablation impact references the percentage increase or decrease in accuracy after the particular electrode signal is replaced with its mean over time. }\label{tbl:s1_ablate}
\end{table}
\newpage

\centering
\includegraphics[scale = 0.5, angle =90]{supp_imgs/ablate_scale}
\begin{table}[h!]
  \centering
  \begin{adjustbox}{max width=\textwidth}
  \begin{tabular}{  c  c  c  }
    
     & ECoG NN & Multimodal NN \\ 
    Pred\_back &
    \begin{minipage}{.55\textwidth}
      \includegraphics[width=\linewidth, ]{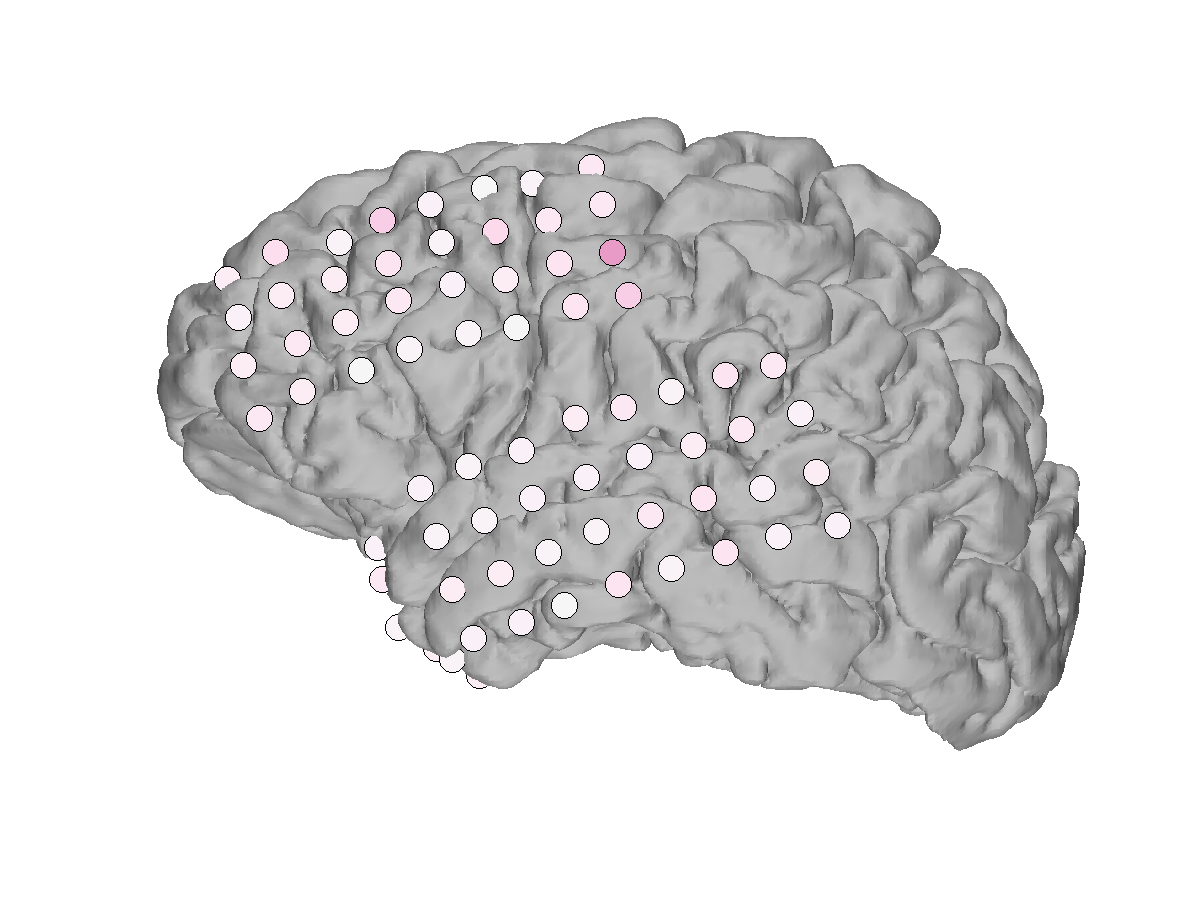}
    \end{minipage}
    & 
    \begin{minipage}{.55\textwidth}
      \includegraphics[width=\linewidth, ]{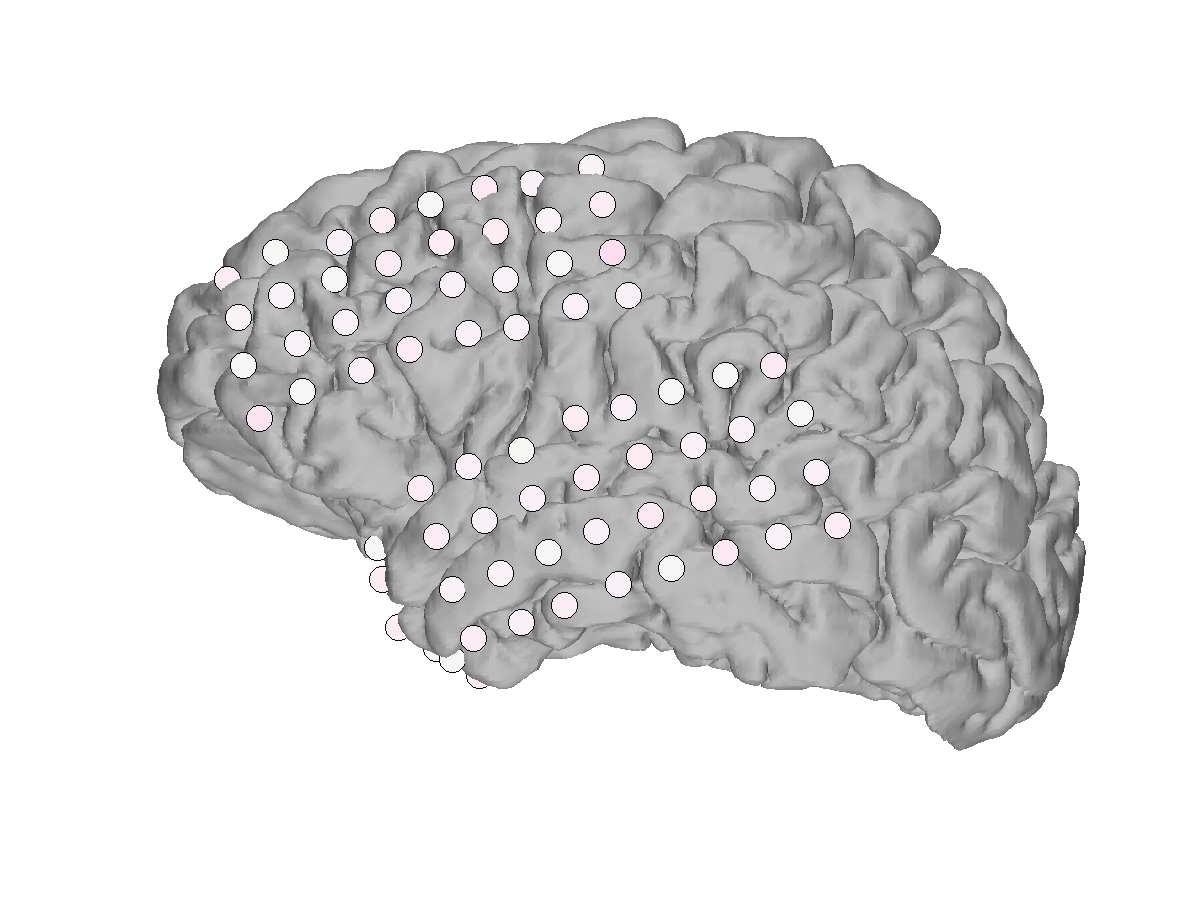}
    \end{minipage} \\ 
    
        Pred & 
    \begin{minipage}{.55\textwidth}
      \includegraphics[width=\linewidth]{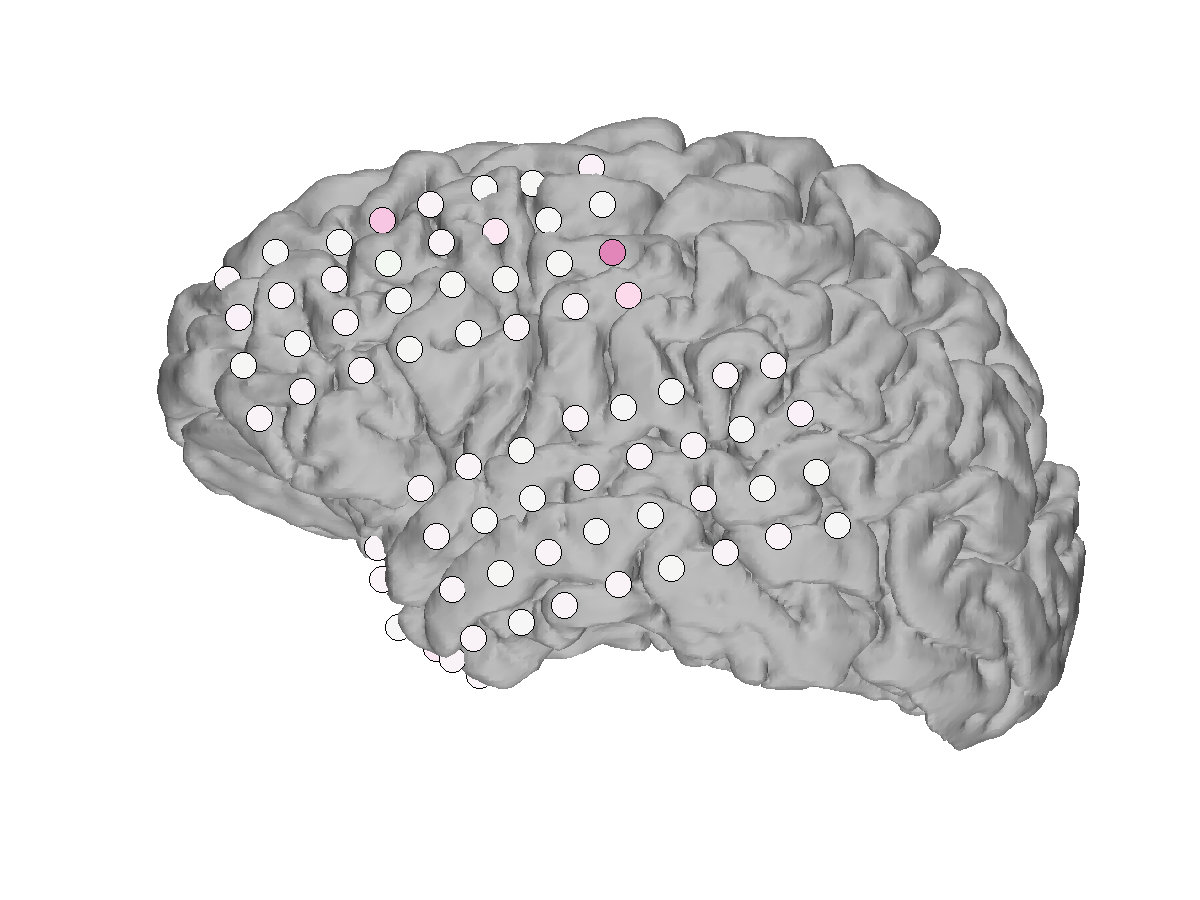}
    \end{minipage}
    &
    \begin{minipage}{.55\textwidth}
      \includegraphics[width=\linewidth]{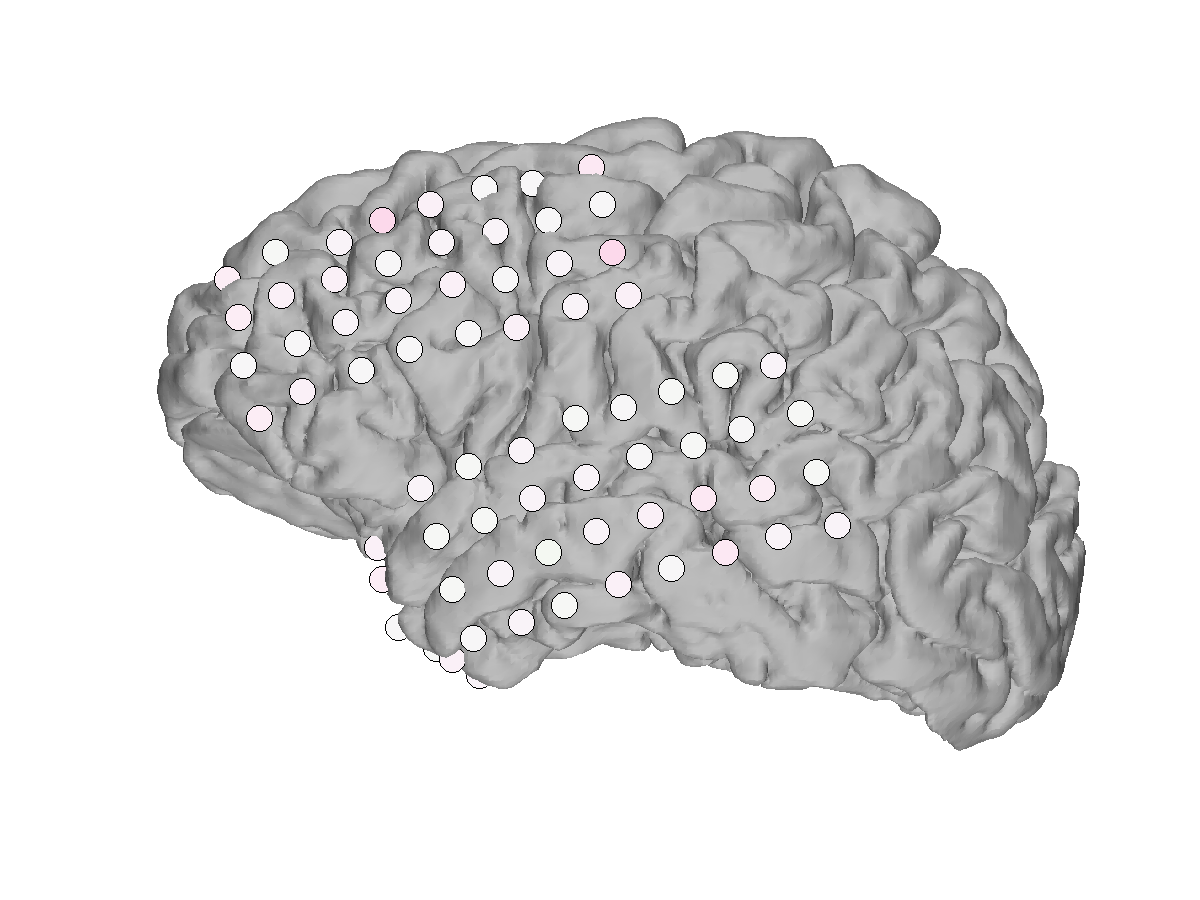}
    \end{minipage} \\ 
    
        Decode &
    \begin{minipage}{.55\textwidth}
      \includegraphics[width=\linewidth]{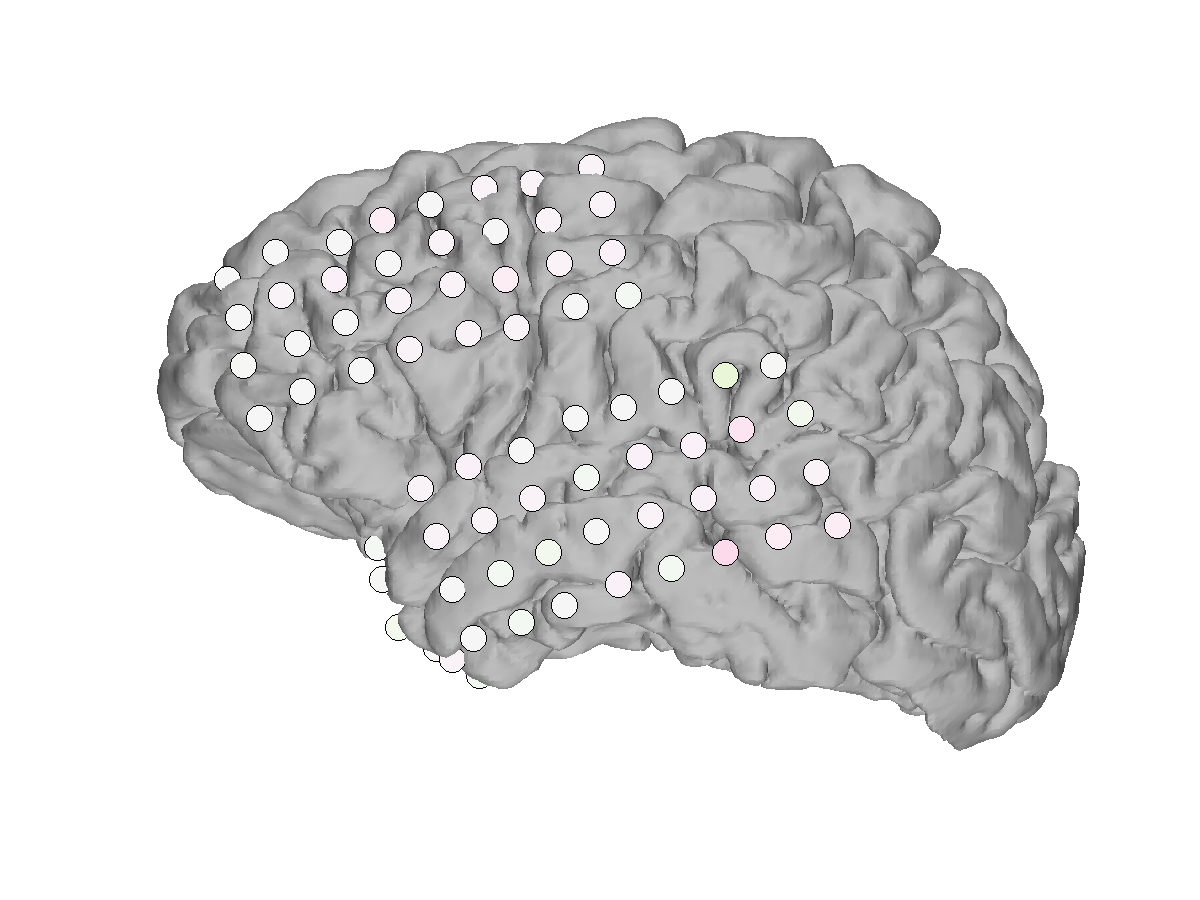}
    \end{minipage}
    &
    \begin{minipage}{.55\textwidth}
      \includegraphics[width=\linewidth]{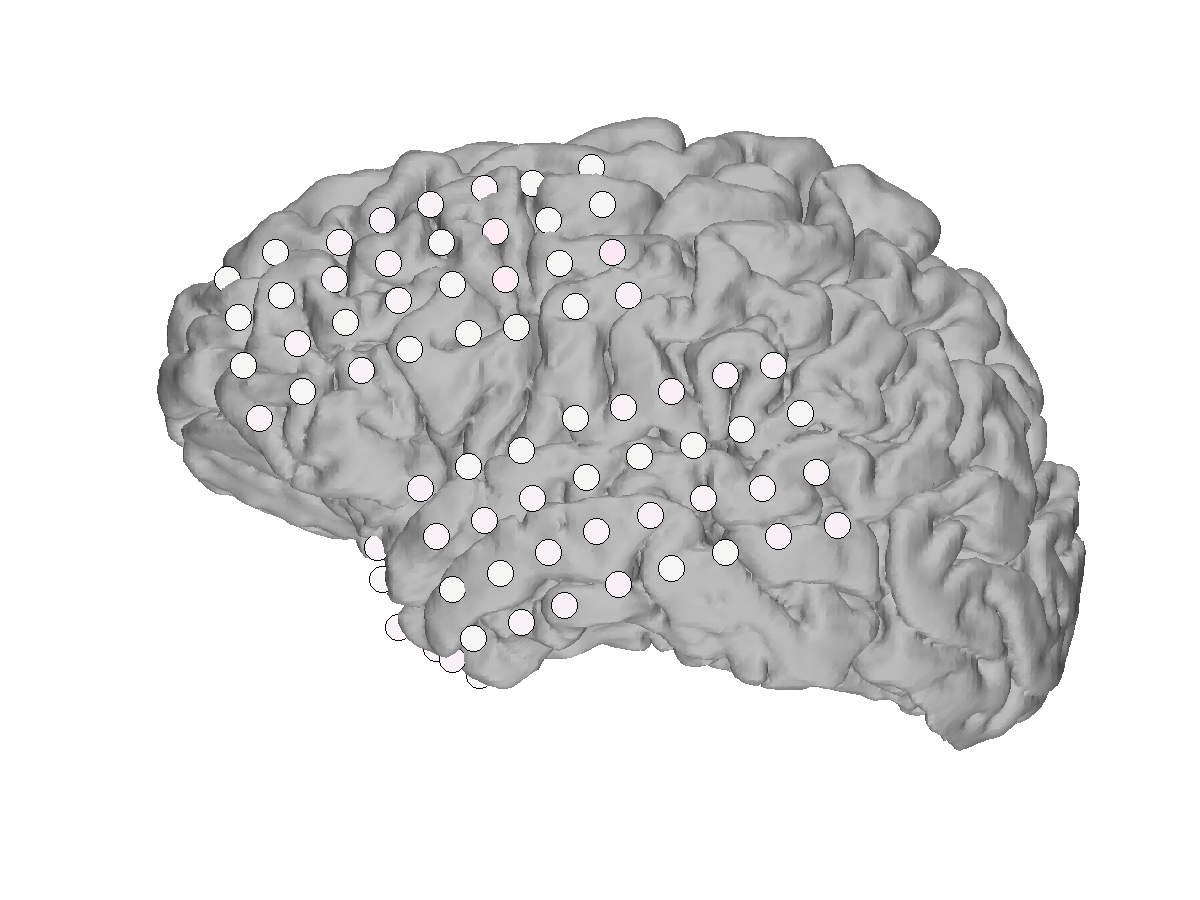}
    \end{minipage} \\ 
  \end{tabular}
  \end{adjustbox}
  \caption*{Figure S3: Subject 2 ablation map. Ablation impact references the percentage increase or decrease in accuracy after the particular electrode signal is replaced with its mean over time.}\label{tbl:s1_ablate}
\end{table}
\newpage

\centering
\includegraphics[scale = 0.5, angle =90]{supp_imgs/ablate_scale}
\begin{table}[h!]
  \centering
  \begin{adjustbox}{max width=\textwidth}
  \begin{tabular}{  c  c  c  }
    
     & ECoG NN & Multimodal NN \\ 
    Pred\_back &
    \begin{minipage}{.55\textwidth}
      \includegraphics[width=\linewidth, ]{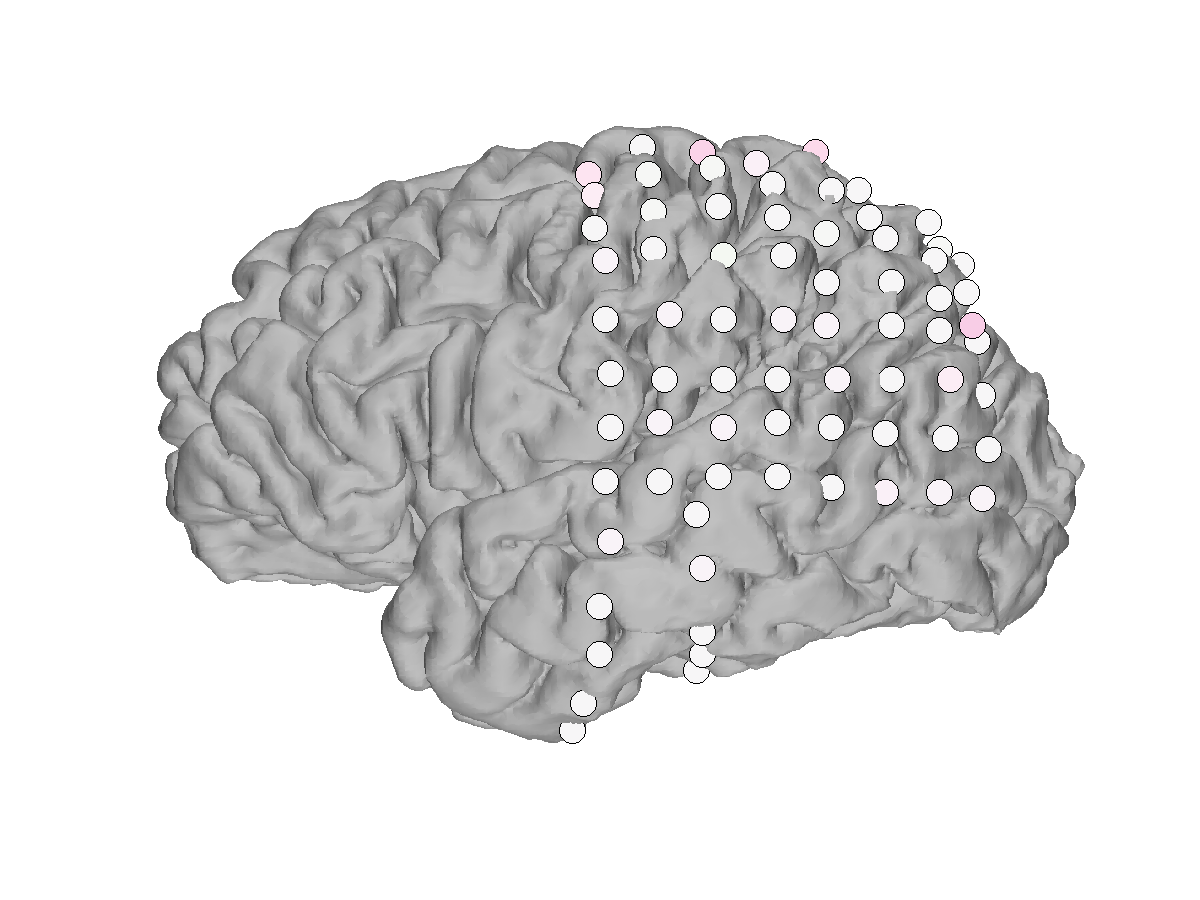}
    \end{minipage}
    & 
    \begin{minipage}{.55\textwidth}
      \includegraphics[width=\linewidth, ]{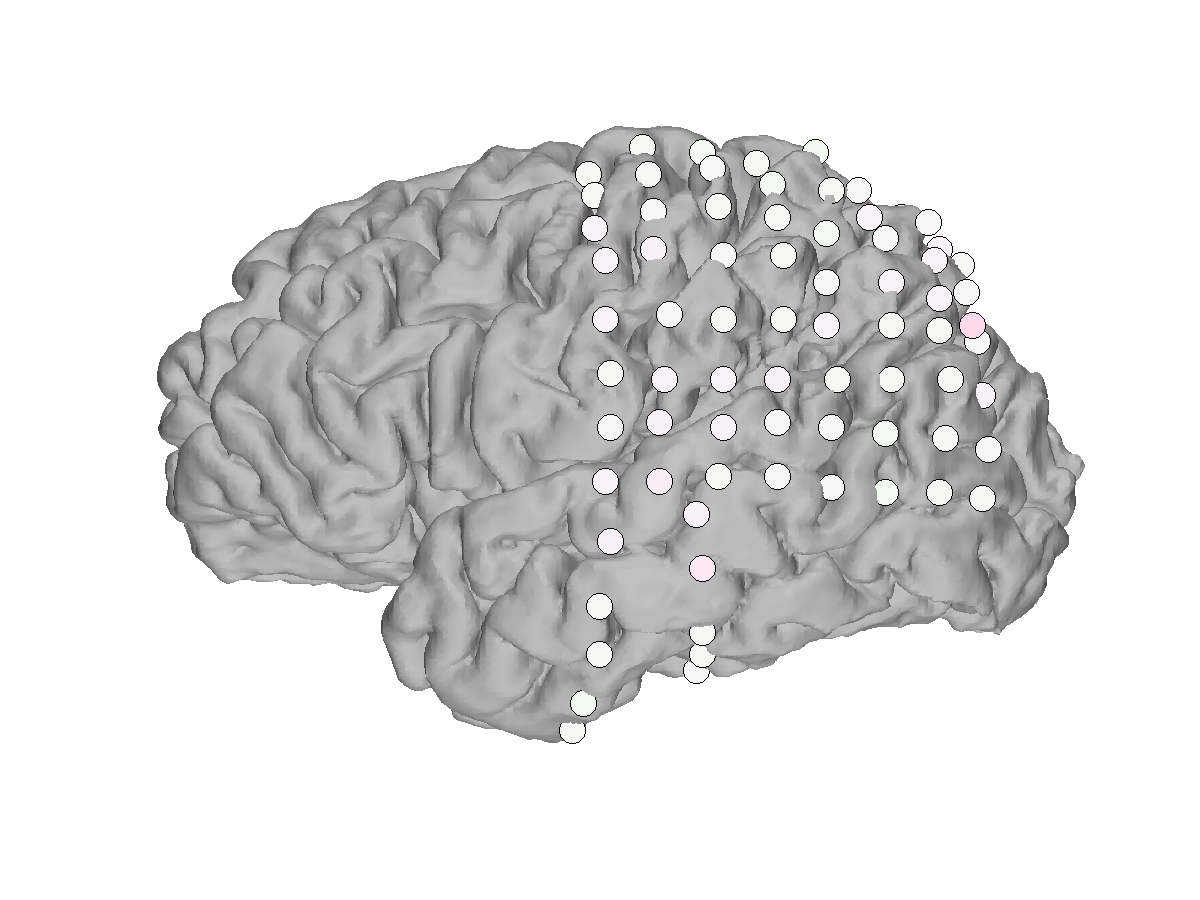}
    \end{minipage} \\ 
    
        Pred & 
    \begin{minipage}{.55\textwidth}
      \includegraphics[width=\linewidth]{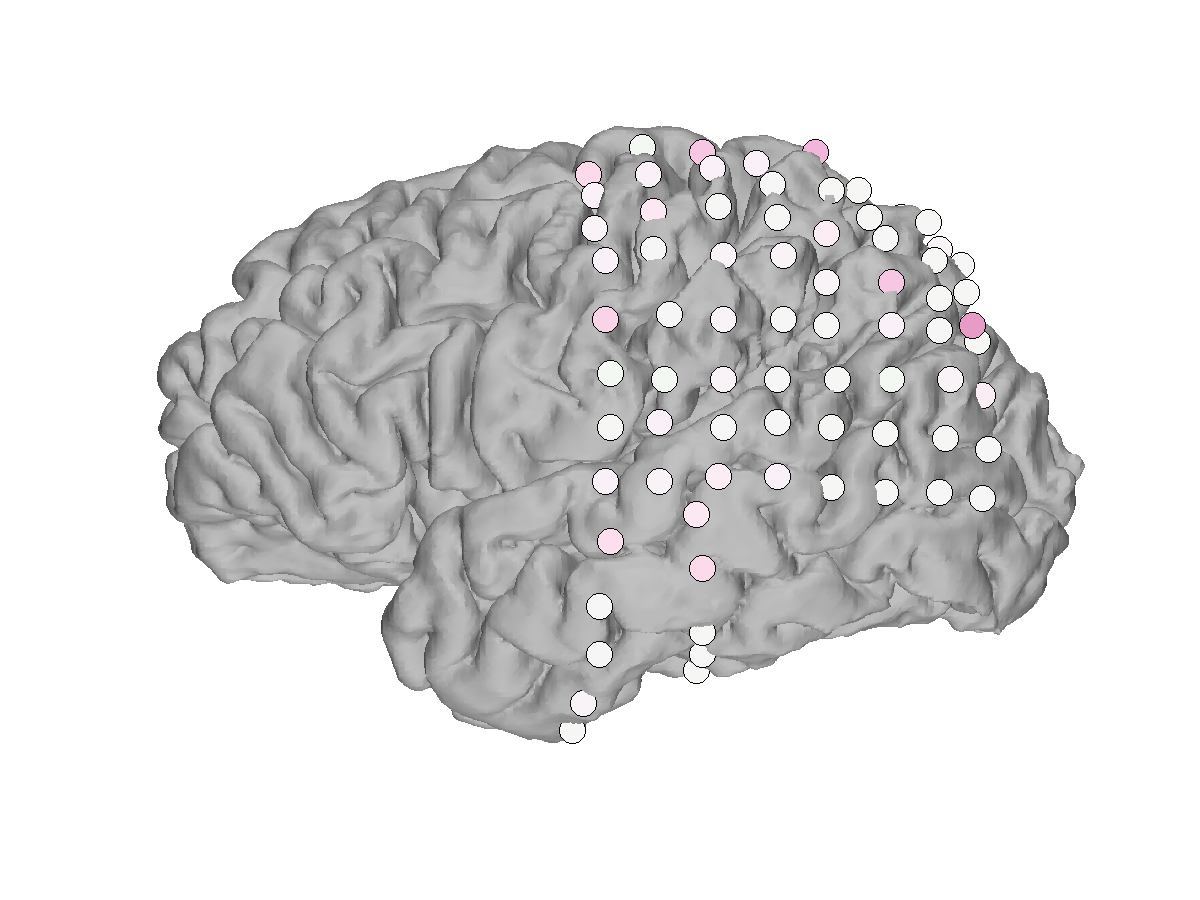}
    \end{minipage}
    &
    \begin{minipage}{.55\textwidth}
      \includegraphics[width=\linewidth]{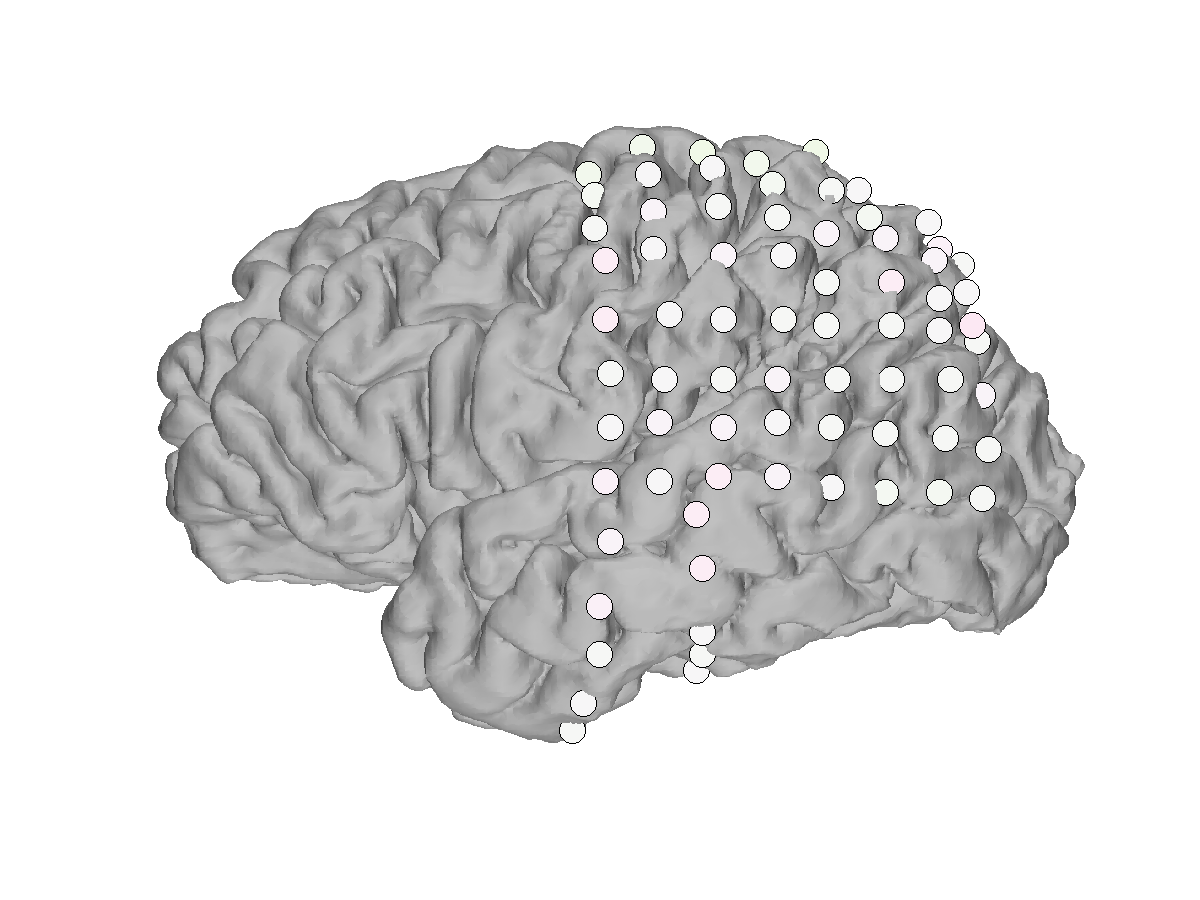}
    \end{minipage} \\ 
    
        Decode &
    \begin{minipage}{.55\textwidth}
      \includegraphics[width=\linewidth]{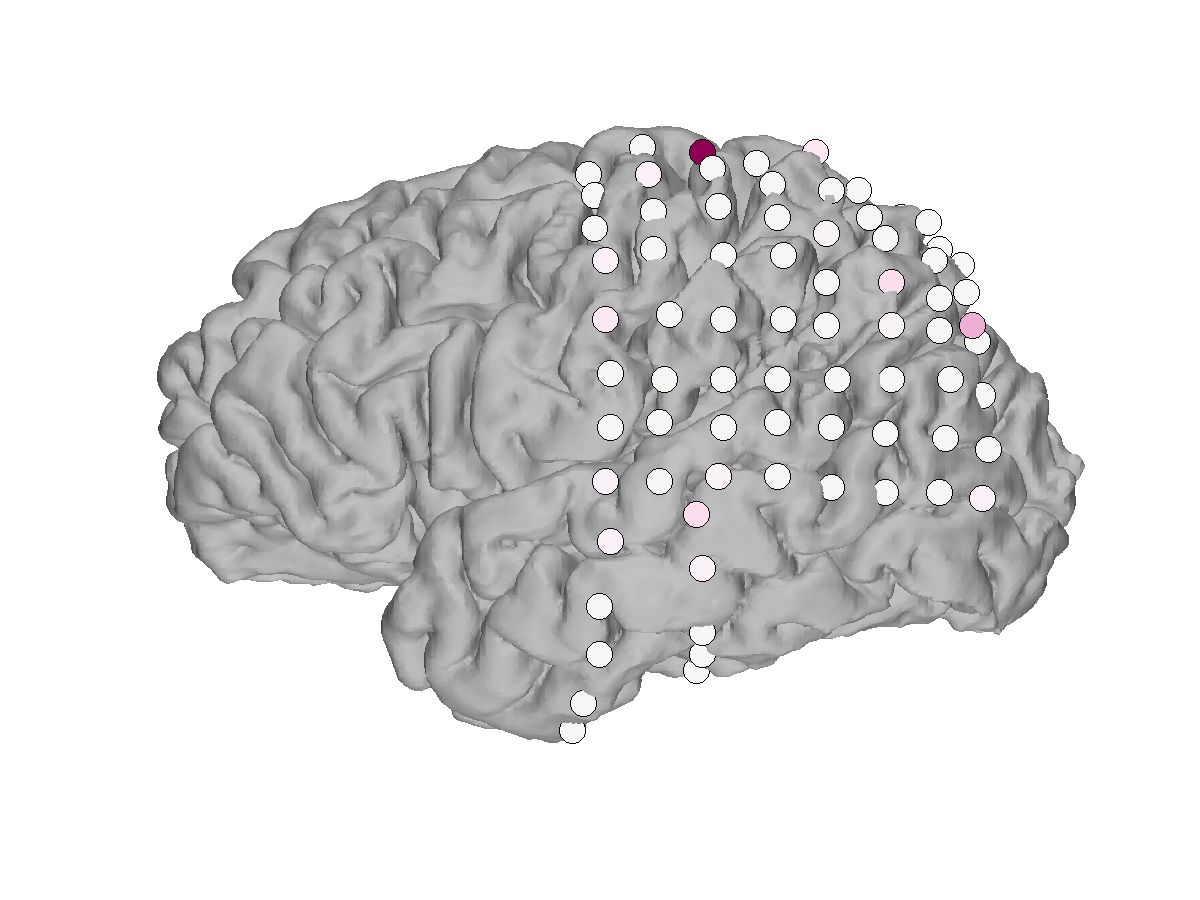}
    \end{minipage}
    &
    \begin{minipage}{.55\textwidth}
      \includegraphics[width=\linewidth]{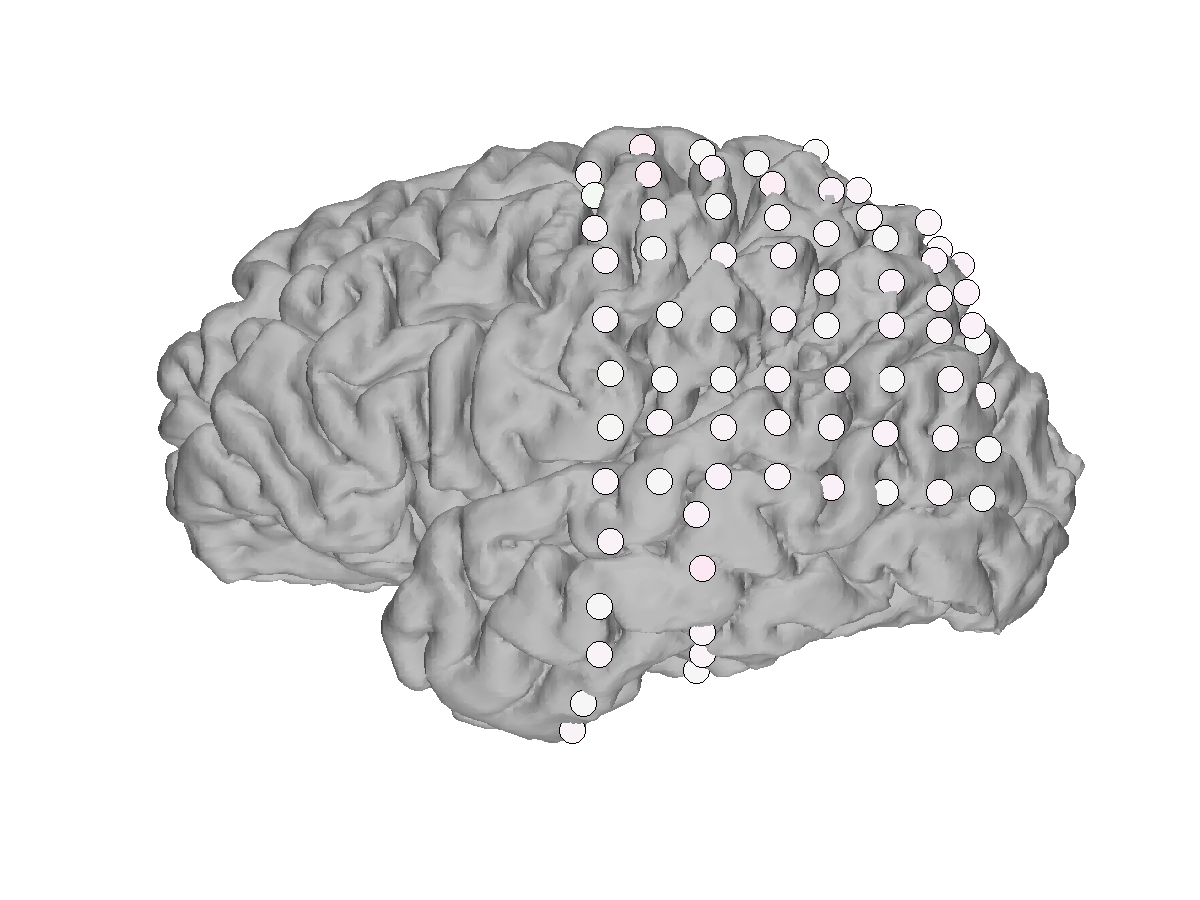}
    \end{minipage} \\ 
  \end{tabular}
  \end{adjustbox}
  \caption*{Figure S4: Subject 3 ablation map. Ablation impact references the percentage increase or decrease in accuracy after the particular electrode signal is replaced with its mean over time.}\label{tbl:s3_ablate}
\end{table}
\newpage

\centering
\includegraphics[scale = 0.5, angle =90]{supp_imgs/ablate_scale}
\begin{table}[h!]
  \centering
  \begin{adjustbox}{max width=\textwidth}
  \begin{tabular}{  c  c  c  }
    
     & ECoG NN & Multimodal NN \\ 
    Pred\_back &
    \begin{minipage}{.55\textwidth}
      \includegraphics[width=\linewidth, ]{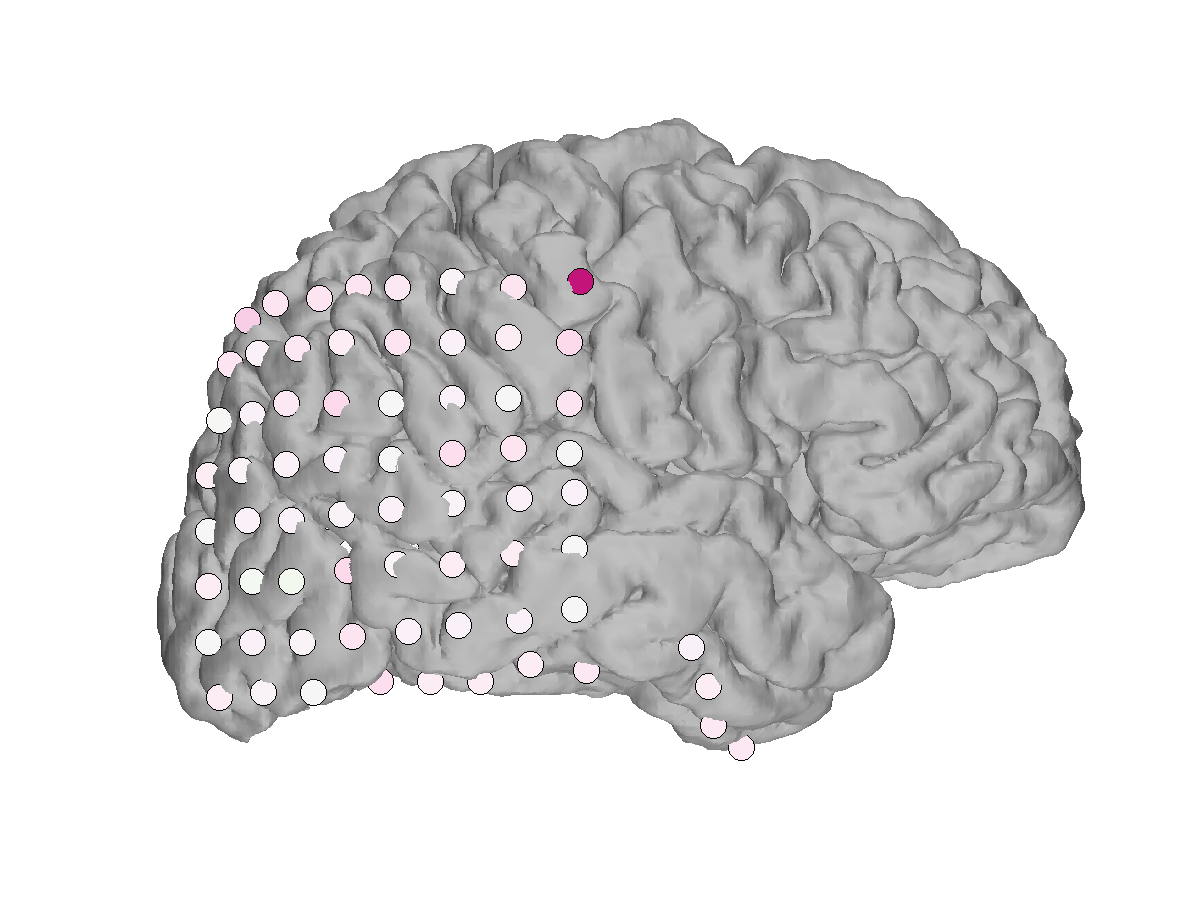}
    \end{minipage}
    & 
    \begin{minipage}{.55\textwidth}
      \includegraphics[width=\linewidth, ]{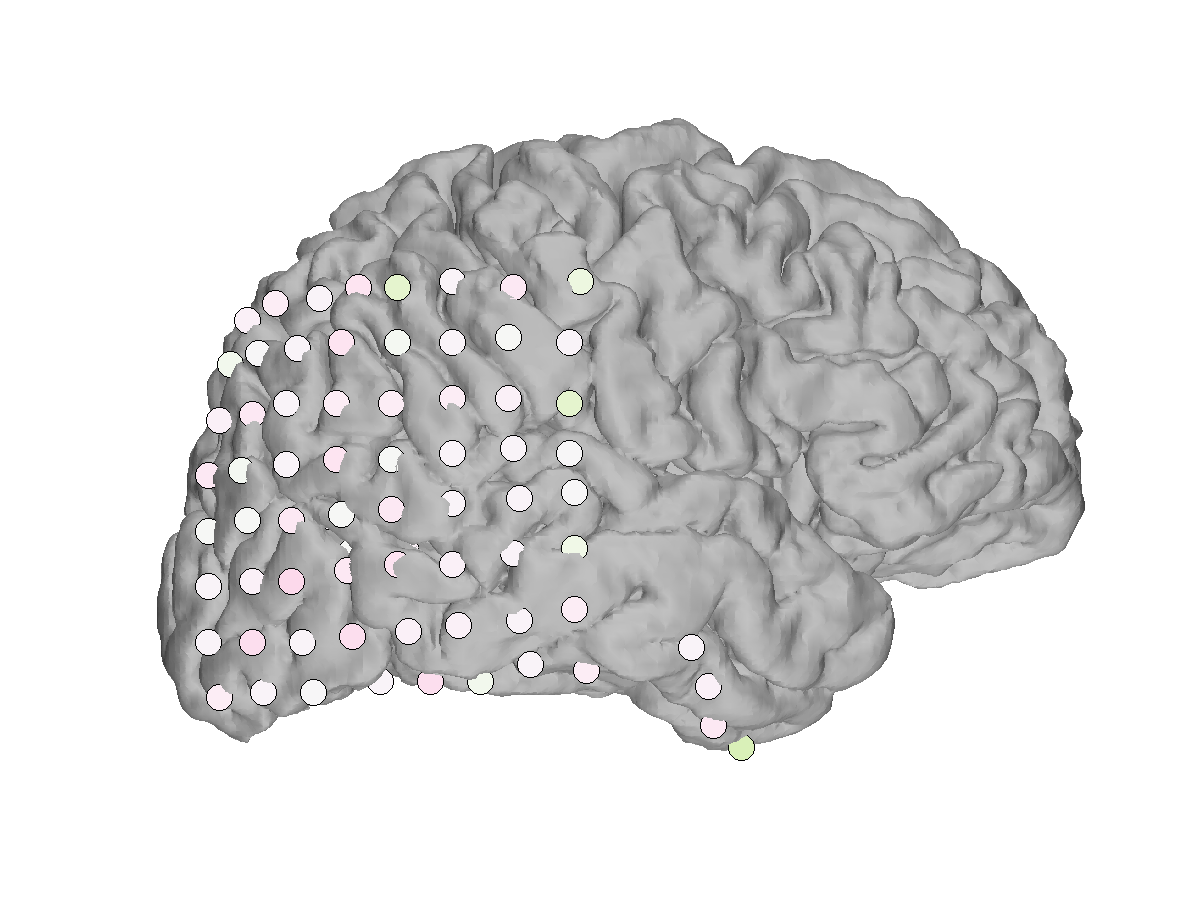}
    \end{minipage} \\ 
    
        Pred & 
    \begin{minipage}{.55\textwidth}
      \includegraphics[width=\linewidth]{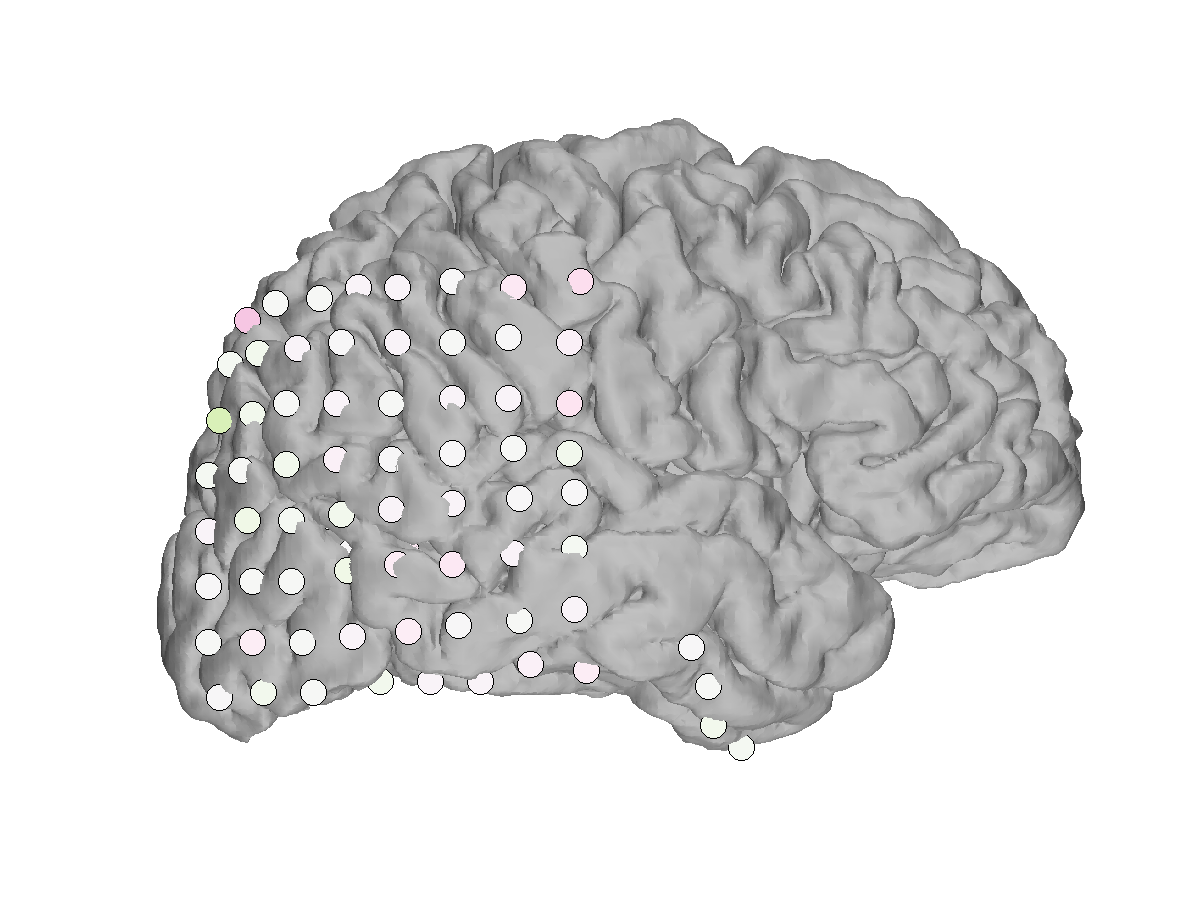}
    \end{minipage}
    &
    \begin{minipage}{.55\textwidth}
      \includegraphics[width=\linewidth]{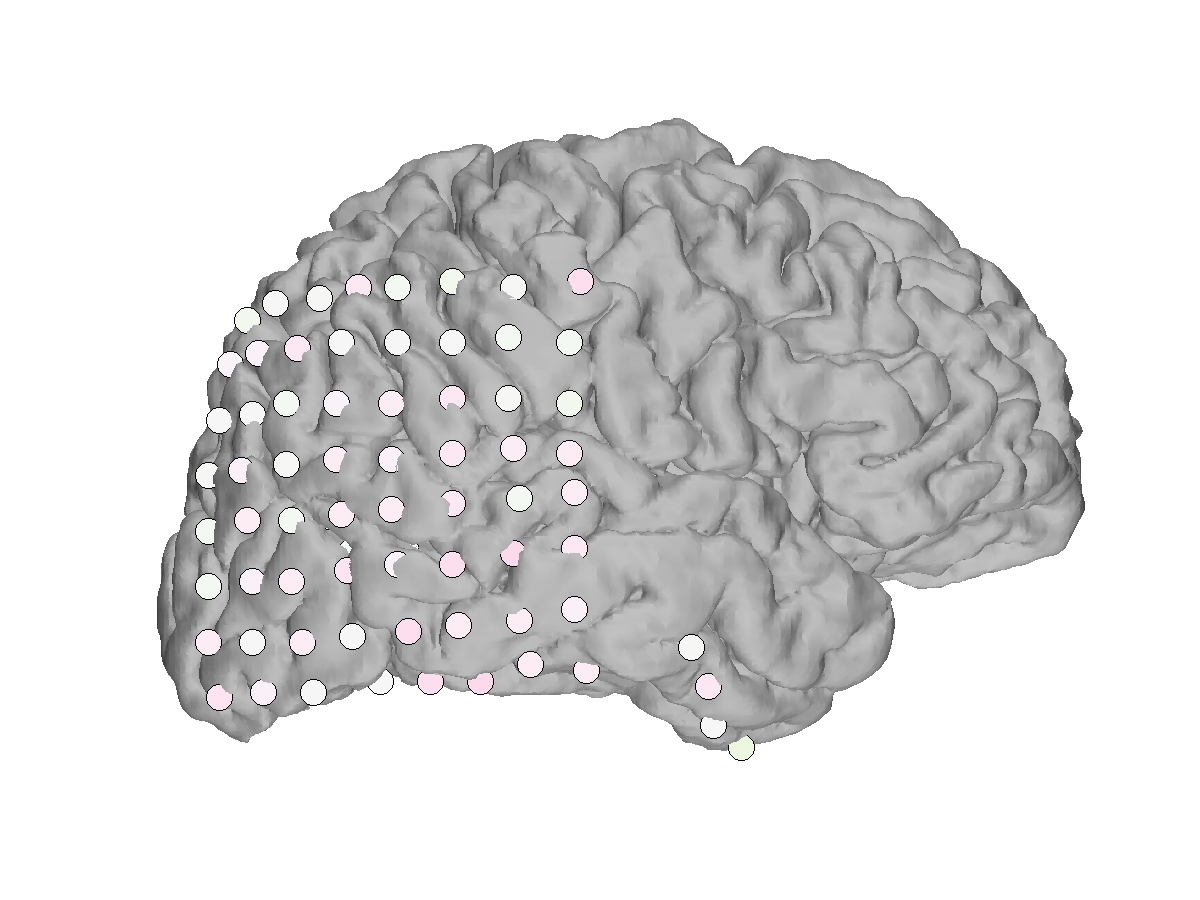}
    \end{minipage} \\ 
    
        Decode &
    \begin{minipage}{.55\textwidth}
      \includegraphics[width=\linewidth]{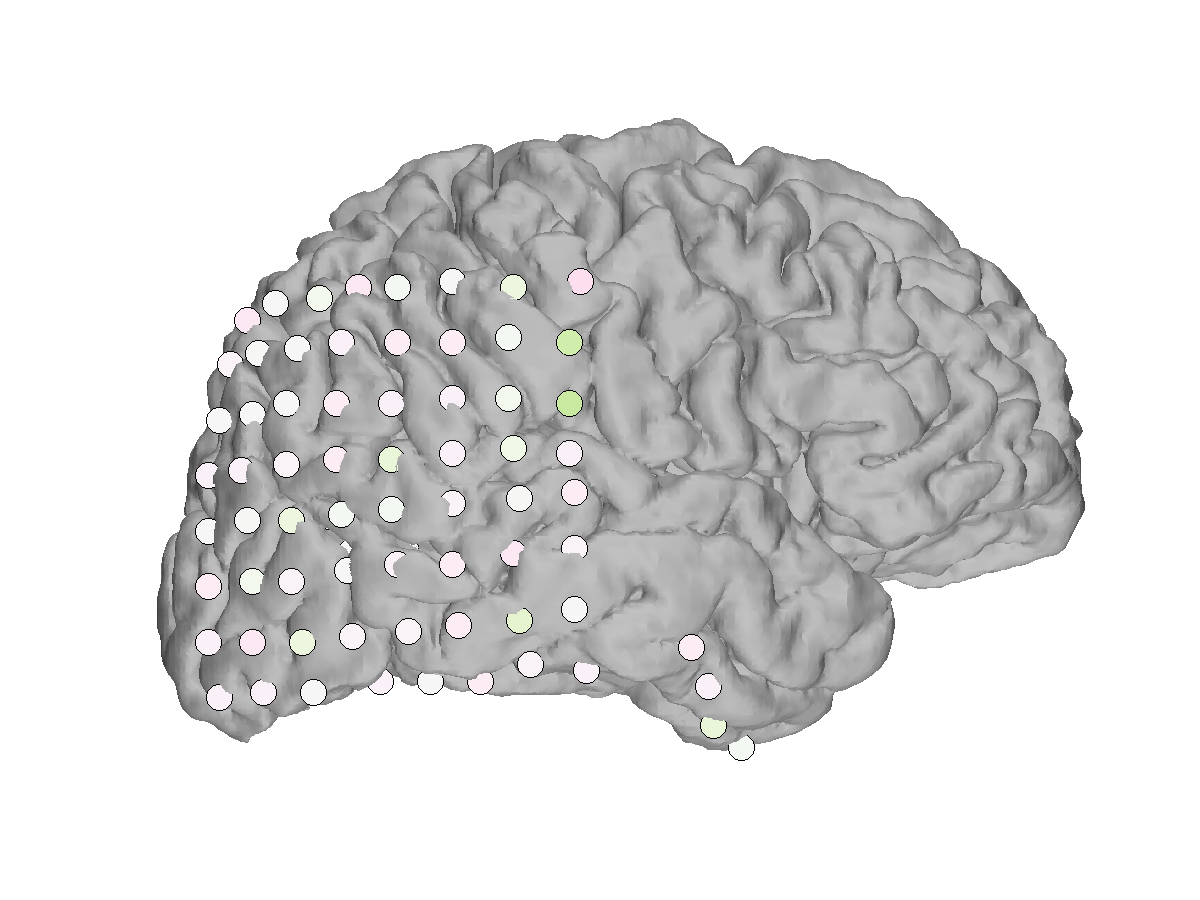}
    \end{minipage}
    &
    \begin{minipage}{.55\textwidth}
      \includegraphics[width=\linewidth]{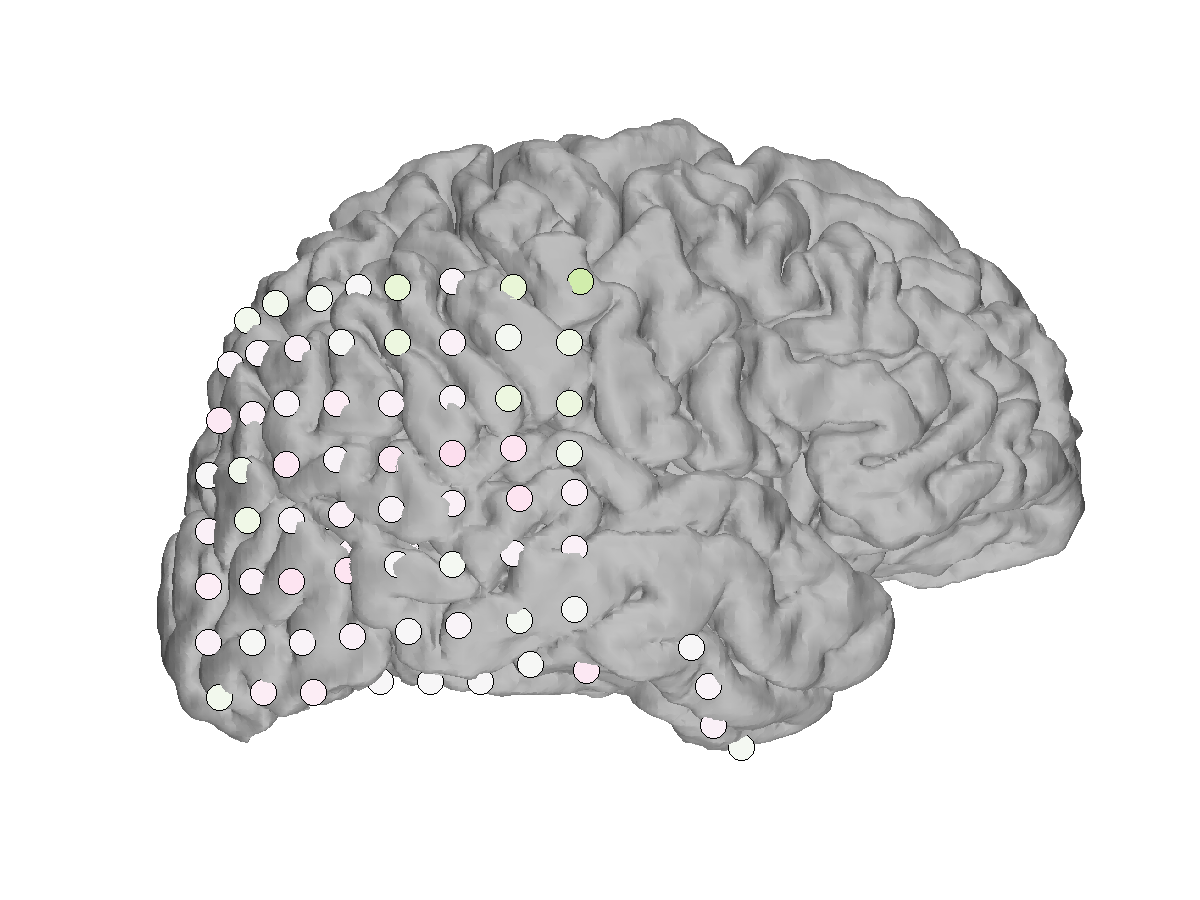}
    \end{minipage} \\ 
  \end{tabular}
  \end{adjustbox}
  \caption*{Figure S5: Subject 4 ablation map. Ablation impact references the percentage increase or decrease in accuracy after the particular electrode signal is replaced with its mean over time.}\label{tbl:s4_ablate}
\end{table}
\newpage